\documentclass[sigconf, 10pt, nonacm]{acmart}
\usepackage{graphicx}
\usepackage[ruled]{algorithm2e}
\usepackage{lipsum}
\usepackage{multicol}
\usepackage{mathtools}

\usepackage{subcaption}
\usepackage{tikz}
\usepackage{pgfplots}
\usetikzlibrary{matrix, calc, fit, positioning, backgrounds}

\tikzstyle{matrix_sty}=[matrix of math nodes, nodes in empty cells, column sep=-1pt, row sep=-1pt, nodes={anchor=center, style={draw=blue!80, inner sep=0pt, line width=1pt, minimum size=25pt}}]

\tikzstyle{dots}=[line width=1pt, dash pattern=on 0pt off 5.75\pgflinewidth, shorten <=5.76\pgflinewidth, line cap=round]
\tikzstyle{ddots}=[line width=1pt, dash pattern=on 0pt off 8.13172798365\pgflinewidth, shorten <=8.13172798365\pgflinewidth, line cap=round]
\tikzstyle{sdots}=[line width=1pt, dash pattern=on 0pt off 8.25\pgflinewidth, shorten <=8.26\pgflinewidth, line cap=round]
\tikzstyle{sddots}=[line width=1pt, dash pattern=on 0pt off 10.7088016796\pgflinewidth, shorten <=10.8\pgflinewidth, line cap=round]
\tikzstyle{ssddots}=[line width=1pt, dash pattern=on 0pt off 11.6672618896\pgflinewidth, shorten <=11.6672618896\pgflinewidth, line cap=round]

\makeatletter
\DeclareRobustCommand{\rvdots}{%
  \vbox{
    \baselineskip4\p@\lineskiplimit\z@
    \kern-\p@
    \hbox{.}\hbox{.}\hbox{.}
  }}
\makeatother

\usepackage{ulem}
\definecolor{darkgreen}{rgb}{0.0, 0.5, 0.0}

\newcommand{\parlabel}[1]{\vspace{0.5em}\noindent\textbf{#1}.}

\newcommand{\Sys}{\textit{UnIT}\xspace}

\graphicspath{{./figs/}}
\AtBeginDocument{%
  }

\setcopyright{acmlicensed}
\copyrightyear{2018}
\acmYear{2018}
\acmDOI{XXXXXXX.XXXXXXX}

\pgfplotsset{compat=1.18}

\begin{document}

\title{\texorpdfstring{\Sys: Scalable \underline{Un}structured \underline{I}nference-\underline{T}ime Pruning for MAC-efficient Neural Inference on MCUs}{}}

\author{Ashe Neth}
\email{aneth@wpi.edu}
\affiliation{
  \institution{Worcester Polytechnic Institute}
  \city{}
  \country{}
}

\author{Sawinder kaur}
\email{sakaur@syr.edu}
\affiliation{
  \institution{Syracuse University}
  \city{}
  \country{}
}

\author{Mohammad Nur Hossain Khan}
\email{mkhan@wpi.edu}
\affiliation{
  \institution{Worcester Polytechnic Institute}
  \city{}
  \country{}
}

\author{Subrata Biswas}
\email{sbiswas@wpi.edu}
\affiliation{
  \institution{Worcester Polytechnic Institute}
  \city{}
  \country{}
}

\author{Asif Salekin}
\email{asif.salekin@asu.edu}
\affiliation{
  \institution{Arizona State University}
  \city{}
  \country{}
}

\author{Bashima Islam}
\email{bislam@wpi.edu}
\affiliation{
  \institution{Worcester Polytechnic Institute}
  \city{}
  \country{}
}



\received{1 July 2025}


\begin{abstract}
Existing pruning methods are typically applied during training or compile time and often rely on structured sparsity. While compatible with low-power microcontrollers (MCUs), structured pruning underutilizes the opportunity for fine-grained efficiency on devices without SIMD support or parallel compute. To address these limitations, we introduce \Sys (\textbf{Un}structured \textbf{I}nference-\textbf{T}ime pruning), a lightweight method that dynamically identifies and skips unnecessary multiply-accumulate (MAC) operations during inference, guided by input-specific activation patterns.
Unlike structured pruning, \Sys embraces irregular sparsity and does not require retraining or hardware specialization. It transforms pruning decisions into lightweight comparisons, replacing multiplications with threshold checks and approximated divisions. \Sys further optimizes compute by reusing threshold computations across multiple connections and applying layer- and group-specific pruning sensitivity. We present three fast, hardware-friendly division approximations tailored to the capabilities of common embedded platforms.
Demonstrated on the MSP430 microcontroller, \Sys achieves 11.02\% to 82.03\% MAC reduction, 27.30\% to 84.19\% faster inference, and 27.33\% to 84.38\% lower energy consumption compared to training-time pruned models, while maintaining accuracy with 0.48-7\%. Under domain shift, \Sys matches or exceeds the accuracy of retrained models while requiring significantly fewer MACs. These results establish unstructured inference-time pruning as a viable and practical solution for efficient, retraining-free deployment of deep neural networks on MCUs.

\end{abstract}

\maketitle
\section{Introduction}
Deploying deep neural networks (DNNs) on low-power microcontroller units (MCUs) is essential for enabling edge intelligence in domains such as healthcare monitoring~\cite{diab2022embedded}, environmental sensing~\cite{gorlatova2010energy, kansal2003environmental}, and industrial automation~\cite{thramboulidis2007soa}. These systems must operate under tight constraints on energy, memory, and compute. In batteryless deployments~\cite{islam2019zygarde, montanari2020eperceptive, farina2024memory, bakar2022protean, gobieski2019intelligence}, the compute budget may be limited to just a few operations per inference cycle. Reducing multiply-accumulate (MAC) operations is therefore critical, as multiplications can be orders of magnitude more expensive than additions. For instance, on the MSP430 MCU, a multiplication takes approximately 77 clock cycles~\cite{MSP430Mul}, while an addition takes only 6~\cite{MSP430}.

Traditional methods for reducing inference cost include quantization~\cite{biswas2025quads, gobieski2019intelligence} and pruning during training or compile time~\cite{shen2022prune, anwar2017structured, kwon2020structured, bakar2022protean}. However, these techniques produce static computation graphs that cannot adapt to input-specific variation or energy fluctuations at runtime. While structured pruning is widely used in high-end accelerators due to its alignment with SIMD and thread-parallel execution models~\cite{yang2018netadapt, cheng2017survey}, it removes entire filters or channels and thus fails to leverage the finer control opportunities offered by single-threaded MCUs with no SIMD support.

In contrast, MCUs are more amenable to unstructured pruning, which retains or skips individual connections. Unlike structured approaches that operate at the granularity of full channels or layers, unstructured pruning provides fine-grained control over which MACs are executed. This can lead to better energy efficiency and accuracy trade-offs, particularly when operating under tight resource budgets. However, implementing unstructured pruning at inference time is challenging due to the overhead of evaluating pruning conditions for each connection and the lack of hardware support for irregular computation.

In addition to hardware constraints, MCUs often operate in dynamic environments where inputs deviate from the training distribution due to user variability, environmental noise, or temporal drift. Pruning strategies designed during training struggle in such settings, as they are optimized for the source domain and lack the flexibility to adapt at inference. Inference-time pruning, by contrast, makes decisions based on the input seen at runtime, allowing the model to skip unnecessary computations while preserving critical ones for out-of-distribution inputs. Prior work on input-aware early exit~\cite{islam2019zygarde, farina2024memory, montanari2020eperceptive} shows that runtime adaptation improves robustness and efficiency on low-power MCUs. However, early exits operate at the layer level—terminating entire inference paths—which limits their granularity and can prematurely discard useful computations. To achieve finer control, we propose input-aware, unstructured pruning at the connection level, enabling the model to retain or remove individual MAC operations based on their runtime utility. This allows deeper layers to be retained when needed, without sacrificing energy efficiency.

To address these challenges, we present \Sys, a novel unstructured inference-time pruning method tailored to low-power MCUs. Instead of pruning weights or activations, \Sys dynamically skips individual MAC operations by checking the significance of each activation-weight pair based on the current input. Each connection corresponds to a scalar MAC, and pruning decisions are made without computing the full multiplication. This is enabled by a key hardware observation: on many MCUs, including the MSP430FR5994, conditional branching requires only 2 to 4 clock cycles, while a multiplication takes approximately 77~\cite{MSP430Mul}. This gap makes connection-level pruning both feasible and efficient.

\Sys introduces a reuse-aware thresholding strategy that avoids computing the full MAC. Rather than multiplying the activation and weight and then comparing the result to a threshold, it pre-divides the layer-wise threshold by either the activation or the weight and compares the remaining term to this result. The choice of reuse direction is based on the layer type: input activations are reused in fully connected layers, so thresholds are computed relative to them; weights are reused in convolutional layers, so they become the basis for comparison. This strategy enables a single threshold division to be reused across multiple MACs, reducing computation and memory overhead without modifying the model architecture.

Even with this optimization, division operations remain a bottleneck since they are nearly as expensive as multiplications on MCUs. Thus, \Sys introduces three lightweight, hardware-specific approximations for the division operation: bit shifting, binary tree search, and bit masking. Each technique is selected based on the numerical format and capabilities of the target device. These estimators allow \Sys to retain its fine-grained pruning capability while ensuring that runtime costs remain negligible, even on resource-constrained embedded platforms.

\Sys is designed for resource-constrained MCUs and integrates seamlessly into existing model compression pipelines. It complements training-time techniques by adding a runtime pruning layer that further improves the trade-off between efficiency and accuracy. Its lightweight, general-purpose logic is especially effective in edge scenarios that require real-time, energy-efficient inference, including deployment on intermittently powered or battery-free devices. Unlike prior inference-time methods, \Sys introduces runtime adaptivity without modifying the model architecture or requiring retraining. This compatibility allows it to layer on top of quantized or training-time pruned models, delivering additional reductions in MAC operations and energy usage with minimal implementation overhead.

\noindent{Our contributions are as follows:}

\begin{itemize}
    \item \textit{MAC-Free, Input-Aware Pruning at Inference Time:} \Sys introduces a new inference-time pruning method that skips individual MAC operations based on the runtime significance of input-weight pairs. Unlike existing approaches that perform the full multiplication to check against a threshold, \Sys avoids the multiplication by reordering the comparison. This enables pruning using only comparisons and a single division, making it well suited for low-power MCUs. To the best of our knowledge, this is the first unstructured pruning strategy that eliminates MACs from threshold evaluation.

    \item \textit{Reuse-Aware Thresholding Across Layer Types:} \Sys selects the control term for pruning based on the reuse pattern in each layer. It uses input activations in fully connected layers and weight values in convolutional layers. This design reduces the number of divisions and allows each computed threshold to be reused across multiple MACs. It supports layer-specific threshold learning and enables generalization across diverse architectures without changing the model structure.

    \item \textit{Hardware-Aware Division Approximation:} To minimize the remaining overhead, \Sys provides three lightweight division approximation algorithms that use bit-level or fixed-point operations. These approximators significantly reduce runtime cost while maintaining pruning accuracy. As a result, \Sys achieves fast, adaptive pruning that meets the performance, memory, and energy constraints of embedded edge devices.

\end{itemize}

We evaluate \Sys on three datasets using the MSP430FR5994 MCU. Our method reduces MAC operations by 11.02--82.03\%, inference time by 27.30--84.19\%\%, and energy consumption by 27.33--84.38\%, while maintaining accuracy within 0.48--7\% of unpruned models. Under domain shift, \Sys matches or outperforms training-time pruned models while requiring 6-10\% fewer MAC operations.

\section{\texorpdfstring{\underline{Un}structured \underline{I}nference-\underline{T}ime Pruning}{}}

\begin{figure*}
\begin{tikzpicture}
    
    \matrix(act1)[matrix_sty, row sep=4pt, style={fill=blue!30}]{
        X_1\\
        X_2\\
        X_3\\
        \node[draw=none]{};\\
        X_m\\
    };
    
    \matrix(weights1)[right=5pt of act1, label=Weights, matrix_sty, row sep=4pt, right delimiter=\}]{
        W_{1,1} & W_{1,2} & W_{1,3} & \node[draw=none]{}; & W_{1,n}\\
        W_{2,1} & W_{2,2} & W_{2,3} & \node[draw=none]{}; & W_{2,n}\\
        W_{3,1} & W_{3,2} & W_{3,3} & \node[draw=none]{}; & W_{3,n}\\
        \node[draw=none]{};\\
        W_{m,1} & W_{m,2} & W_{m,3} & \node[draw=none]{}; & W_{m,n}\\
    };

    \matrix(alg1)[right=of weights1, matrix_sty, row sep=4pt, column sep=4pt]{
        X_i & \node[draw=none]{}; & \node(alg1-1-3)[shape=circle, draw=red!80]{\frac{T}{|X_i|}}; & \node[draw=none]{}; & \node(alg1-1-5)[draw=violet!80]{t_i};\\
        \node[draw=none, minimum size = 10pt]{};\\
        W_{i,1} & W_{i,2} & W_{i,3} & \node[draw=none]{}; & W_{i,n}\\
        \node(alg1-4-1)[draw=none]{}; & \node(alg1-4-2)[draw=none]{}; & \node(alg1-4-3)[draw=none]{}; & \node(alg1-4-4)[draw=none]{};\\
        \node(alg1-5-1)[draw=none, minimum size = 10pt]{\texttt{No}}; & \node[draw=none, minimum size = 10pt]{}; & \node[draw=none, minimum size = 10pt]{}; & \node(alg1-5-4)[draw=none, minimum size = 10pt]{\texttt{Yes}};\\
        \node(alg1-6-1)[draw=none]{}; & \node[draw=none]{}; & O_j & \node(alg1-6-4)[draw=none]{+=}; & \node(alg1-6-5)[draw=none]{};\\
    };

    \node(check)[draw=none, anchor=center]at(alg1-4-3){$|W_{i,j}|>t_i$};
    \node(skip)[draw=none, anchor=center]at(alg1-6-1){Skip};
    \node(mul)[draw=none, anchor=center]at(alg1-6-5){$X_iW_{i,j}$};

    \node[label={[label distance=-30pt,rotate=90]right:Previous Layer Output}]at([xshift=-15pt]act1-3-1.west){};
    
    \draw[sdots](act1-3-1)--(act1-5-1);
    
    \draw[dots](weights1-1-3)--(weights1-1-5);
    \draw[dots](weights1-2-3)--(weights1-2-5);
    \draw[dots](weights1-3-3)--(weights1-3-5);
    \draw[dots](weights1-5-3)--(weights1-5-5);
    
    \draw[sdots](weights1-3-1)--(weights1-5-1);
    \draw[sdots](weights1-3-2)--(weights1-5-2);
    \draw[sdots](weights1-3-3)--(weights1-5-3);
    \draw[sdots](weights1-3-5)--(weights1-5-5);

    \draw[sddots](weights1-3-3)--(weights1-5-5);
    
    \node[draw=black!20!green, inner sep=1pt, line width=1pt, fit={(act1-1-1) (weights1-1-5)}]{};
    \node[draw=black!20!green, inner sep=1pt, line width=1pt, fit={(act1-2-1) (weights1-2-5)}]{};
    \node[draw=black!20!green, inner sep=1pt, line width=1pt, fit={(act1-3-1) (weights1-3-5)}]{};
    \node[draw=black!20!green, inner sep=1pt, line width=1pt, fit={(act1-5-1) (weights1-5-5)}]{};

    \draw[sdots](alg1-3-3)--(alg1-3-5);

    \draw[line width=2pt, ->](alg1-1-1)--(alg1-1-3);
    \draw[line width=2pt, ->](alg1-1-3)--(alg1-1-5);
    
    \draw[line width=2pt, shorten <=0.5\pgflinewidth, ->](check)-|(alg1-4-1)-|(alg1-5-1);
    \draw[line width=2pt, shorten <=0.5\pgflinewidth, ->](alg1-5-1)--(skip);
    \draw[line width=2pt, shorten <=0.5\pgflinewidth, ->](check)-|(alg1-4-4)-|(alg1-5-4);
    \draw[line width=2pt, shorten <=0.5\pgflinewidth, ->](alg1-5-4)--([yshift=-5pt]alg1-6-4.north);
    
    \node(weight-foc)[draw=black!20!green, inner sep=1pt, line width=1pt, fit={(alg1-3-1) (alg1-3-5)}]{};

    \draw[thick, decorate, decoration={brace,mirror}]([xshift=-2pt,yshift=-2pt]weight-foc.south west)--([xshift=2pt,yshift=-2pt]weight-foc.south east){};
\end{tikzpicture}
\caption{In a fully connected network, traditional pruning applies a fixed mask to zero out inputs regardless of their influence. \Sys introduces input-aware pruning by leveraging the reuse of inputs across multiple output connections. For each input $X_i$, a dynamic threshold $t_i = \frac{T}{|X_i|}$ is computed. Because each $X_i$ contributes to all downstream outputs, this formulation enables more efficient pruning decisions: if $|W_{i,j}| > t_i$, the corresponding MAC is computed; otherwise, it is skipped. This approach preserves high-impact activations while reducing redundant computation.}
\label{fig:linear}
\end{figure*}
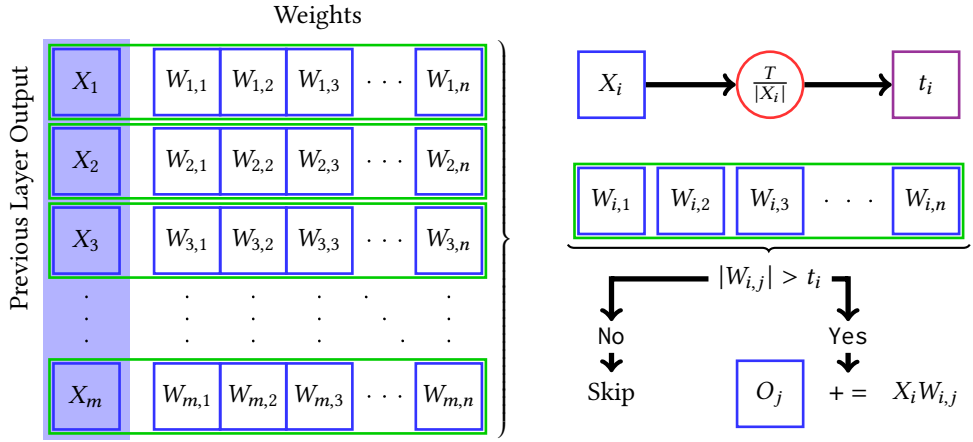

\noindent\textbf{Objective.}  
The objective of this work is to develop a fine-grained, unstructured inference-time pruning method that minimizes multiply-accumulate (MAC) operations on low-power microcontrollers (MCUs) without requiring retraining or changes to model architecture. Our goal is to reduce computational and energy overhead through input-adaptive pruning while maintaining accuracy across varying tasks and data distributions.

\vspace{0.5em}
\noindent\textbf{Key Questions. }
How can we enable unstructured, input-aware inference-time pruning that is efficient enough to run on memory- and energy-constrained MCUs, while avoiding expensive operations such as multiplication and division?

\vspace{0.5em}
\noindent\textbf{Building Blocks for \Sys.}
\Sys addresses this challenge by enabling MAC-free pruning decisions through two core components: Connection-Level Pruning via Reuse-Aware Thresholding and Fast Division Approximation. A connection refers to a scalar multiply-accumulate (MAC) operation between an input activation and its corresponding weight. Instead of performing this multiplication to evaluate pruning decisions, \Sys reformulates the condition into a comparison that avoids direct MAC computation. This is achieved by dividing a layer-wise threshold by a control term and reusing the result across multiple connections. The control term is selected based on the layer’s reuse structure: in linear layers, reused input activations serve as control terms, while in convolutional layers, it is the shared kernel weights. This reuse-aware thresholding strategy reduces the number of division operations required, supports input-adaptive pruning, and preserves the original model architecture. To further reduce computational cost, \Sys incorporates Fast Division Approximation, a hardware-friendly technique that substitutes expensive division with efficient bit-level or fixed-point operations, making it well-suited for deployment on low-power microcontrollers.

\subsection{MAC-Free Connection Pruning and Reuse-Aware Thresholding}
\label{sec:thresholding}

Traditional unstructured inference-time pruning~\cite{kurtz2020inducing} typically involves computing the product of an input activation and its corresponding weight, then zeroing out the result if it falls below a fixed threshold. While this avoids accumulating insignificant outputs, it still requires a full MAC operation before any pruning decision is made, which is costly on low-power microcontrollers (MCUs).

\Sys avoids this overhead by pruning at the connection level, where each connection refers to a scalar MAC operation between an input activation and a weight. Instead of calculating the full product, \Sys reformulates the pruning condition to avoid multiplication altogether. This is done by dividing a layer-wise threshold by a reused control term and comparing the result to the remaining MAC operand:

\begin{equation}
    |\mathbf{X}_{i,l} \cdot \mathbf{W}_{l,j}| \leq T \quad \Leftrightarrow \quad |\mathbf{Z}| \leq \frac{T}{|C|}
\end{equation}

Here, $\mathbf{Z}$ is either the activation or the weight, and $C$ is the corresponding control term, selected based on which operand is more frequently reused. This enables a single division to be reused across multiple MAC decisions.

\noindent\textbf{Linear Layers.} 
Fully connected (linear) layers use large weight matrices where each weight $\mathbf{W}_{l,j}$ is involved in only a single MAC. However, each activation $\mathbf{X}_{i,l}$ is typically reused across multiple output neurons. Thus, we reverse the normalization in Equation~\ref{eq:input}, computing thresholds relative to the activation. This allows the same threshold to be applied across a row of weights in a dense layer, amortizing computation across multiple MACs. Figure~\ref{fig:linear} illustrates this reuse pattern.

\begin{equation}
    \Hat{\mathbf{W}}_{l,j} =
\begin{dcases}
    0 & \textnormal{if }|\mathbf{W}_{l,j}| \leq \Bar{\mathbf{X}}_{i, l} \\
    \mathbf{W}_{l,j}& \textnormal{otherwise}
\end{dcases}
\quad
\text{where } \Bar{\mathbf{X}}_{i, l} = \frac{T}{|\mathbf{X}_{i, l}|}
\label{eq:input}
\end{equation}

\begin{figure*}
    \centering
    \subfloat[Train-time pruning removes the same weight connections for all inputs by applying a fixed binary mask, regardless of input variability.\label{standard}]{\includegraphics[width=.42\textwidth]{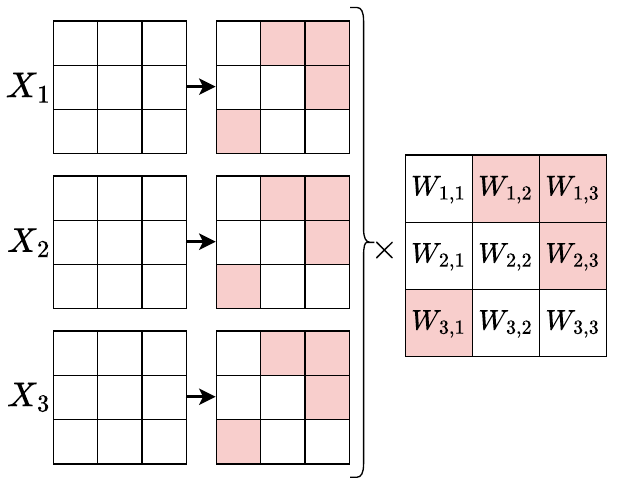}}
    \hspace{1em}
    \subfloat[  \Sys enables dynamic, input-aware pruning by comparing each input $X_i$ to a connection-specific threshold $\tau_{i,j} = \frac{T}{|W_{i,j}|}$. Since the layer-level threshold $T$ is fixed and weights are reused across multiple connections in convolution, these $\tau_{i,j}$ values can be reused efficiently across multiple connections.
     \label{s}]{\includegraphics[width=.55\textwidth]{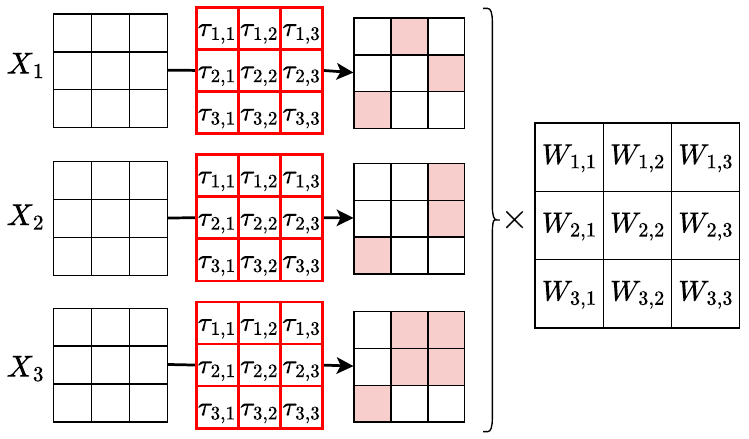}}
    \caption{Comparison between train-time and inference-time pruning strategies. While train-time pruning applies a fixed binary mask regardless of input variability, \Sys enables dynamic, input-aware pruning by adapting thresholds to each activation-weight pair.}
    \label{fig:conv}
\end{figure*}

\noindent\textbf{Convolutional Layers.} 
In convolutional layers, weights are reused across spatial positions, while activations are unique per location. \Sys therefore selects weights, each weight $\mathbf{W}_{l,j}$,  as control terms (Equation~\ref{eq:weight}). This approach (shown in Figure~\ref{fig:conv}) enables threshold reuse across many MACs involving the same weight, amortizing the cost of computing $\dfrac{T}{|\mathbf{W}_{l,j}|}$.

\begin{equation}
    \Hat{\mathbf{X}}_{i,l} =
\begin{dcases}
    0 & \textnormal{if }|\mathbf{X}_{i,l}| \leq \Bar{\mathbf{W}}_{l,j} \\
    \mathbf{X}_{i,l}& \textnormal{otherwise}
\end{dcases}
\quad
\text{where } \Bar{\mathbf{W}}_{l,j} = \frac{T}{|\mathbf{W}_{l,j}|}
\label{eq:weight}
\end{equation}

\noindent\textbf{Adaptive Threshold Calibration.}
Since activation and weight distributions vary across layers, using a single global threshold leads to inconsistent pruning. \Sys includes a one-time calibration phase that collects statistics of activation-weight products from a held-out batch to compute per-layer thresholds using a fixed percentile (e.g., 20th). These thresholds are stored as constants in the final model binary and require no runtime computation or memory.

\noindent\textbf{Fine-Grained and Deterministic Pruning.}
\Sys also supports optional group-wise thresholding within layers for more nuanced pruning. This strategy allows one division to guide multiple MAC decisions, reducing hardware cost while preserving flexibility. The entire thresholding approach remains deterministic, interpretable, and hardware-efficient and avoids the unpredictability of schedule-based or reinforcement learning-based pruning strategies.

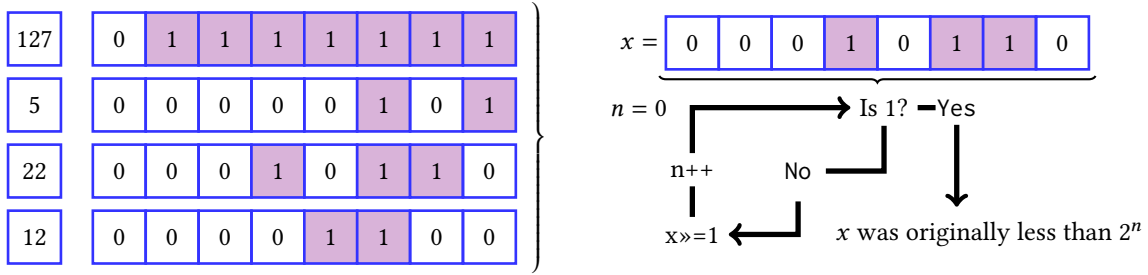
\begin{figure*}
\begin{tikzpicture}
        \matrix(in_values)[matrix_sty, row sep=4pt, nodes={minimum size=20pt}]{
           127\\
           5\\
           22\\
           12\\
        };
        
        \matrix(in_bits)[right=4pt of in_values, matrix_sty, row sep=4pt, nodes={minimum size=20pt}, right delimiter=\}]{
           0 & \node[fill=violet!30]{1}; & \node[fill=violet!30]{1}; & \node[fill=violet!30]{1}; & \node[fill=violet!30]{1}; & \node[fill=violet!30]{1}; & \node[fill=violet!30]{1}; & \node[fill=violet!30]{1};\\
           0 & 0 & 0 & 0 & 0 & \node[fill=violet!30]{1}; & 0 & \node[fill=violet!30]{1};\\
           0 & 0 & 0 & \node[fill=violet!30]{1}; & 0 & \node[fill=violet!30]{1}; & \node[fill=violet!30]{1}; & 0\\
           0 & 0 & 0 & 0 & \node[fill=violet!30]{1}; & \node[fill=violet!30]{1}; & 0 & 0\\
        };

        \matrix(alg)[right=of in_bits, matrix_sty, row sep=4pt, nodes={minimum size=20pt}]{
           \node[draw=none]{x=}; & 0 & 0 & 0 & \node[fill=violet!30]{1}; & 0 & \node[fill=violet!30]{1}; & \node[fill=violet!30]{1}; & 0\\
           \node[draw=none]{n=0}; & \node(loop)[draw=none]{}; & \node[draw=none]{}; & \node[draw=none]{}; & \node(alg-2-5)[draw=none]{}; & \node[draw=none]{}; & \node(alg-yes)[draw=none, minimum size = 10pt]{\texttt{Yes}};\\
           \node[draw=none]{}; & \node(inc)[draw=none]{}; & \node[draw=none]{}; & \node(alg-no)[draw=none]{\texttt{No}}; & \node(mid)[draw=none]{}; & \node[draw=none]{};\\
           \node[draw=none]{}; & \node(alg-5-1)[draw=none]{}; & \node[draw=none]{}; & \node[draw=none]{}; \node(alg-6-4)[draw=none]{}; & \node[draw=none]{}; & \node[draw=none]{}; & \node(end)[draw=none]{};\\
        };

        \node(check)[draw=none, anchor=center]at([xshift=12.5pt]alg-2-5.center){Is 1?};
        \node(incr)[draw=none, minimum size = 10pt, anchor=center]at(inc){n++};
        \node(shift)[draw=none, minimum size = 10pt, anchor=center]at(alg-5-1){x>>=1};
        \node(ending)[draw=none, anchor=center]at([xshift=12.5pt]end.center){$x$ was originally less than $2^n$};

        \draw[line width=2pt, shorten <=0.5\pgflinewidth, ->](check)--([xshift=12.5pt]mid.center)--(alg-no)--(alg-6-4.center)--(shift);
        \draw[line width=2pt, shorten <=0.5\pgflinewidth, ->](shift)--(incr)--(loop.center)--(check);
        \draw[line width=2pt, shorten <=0.5\pgflinewidth, ->](check)--(alg-yes)--(end);
        
        \draw[thick, decorate, decoration={brace,mirror}]([xshift=-2pt,yshift=-2pt]alg-1-2.south west)--([xshift=2pt,yshift=-2pt]alg-1-9.south east){};
\end{tikzpicture}

\caption{Bit shifting can approximate the order of magnitude of any integer or fixed-point number $x$. By repeatedly shifting $x$ to the right and counting the number of shifts $n$ until the most significant bit becomes 1, we estimate the smallest $n$ such that $x < 2^n$. This requires at most $\omega$ shifts, where $\omega$ is the processor’s word size.}
\label{fig:shift}
\end{figure*}

\subsection{Fast Division Approximation}
While \Sys avoids most multiplications by comparing a single activation or weight to a reusable threshold ratio (Section~\ref{sec:thresholding}), the division used to compute this threshold, which is shared across multiple connections, can still be costly on embedded platforms. To address this, \Sys introduces three lightweight and hardware-specific division approximations: \textit{Bit Shifting}, \textit{Binary Tree Search}, and \textit{Bit Masking}. These methods allow pruning thresholds to be estimated with minimal latency, enabling deployment on a wide range of microcontrollers, from ultra-low-power fixed-point devices (e.g., MSP430FR5994) to floating-point accelerators (e.g., MAX78000). Each approximation exploits numeric representations specific to its target architecture to minimize instruction count, energy, and compute time.

\vspace{0.5em}
\noindent\textbf{Bit Shifting (Fixed-Point / Integer Devices).}
Bit shifting approximates division by computing the exponent of a value in its binary form. For a scalar input $\mathbf{X}$ and threshold $\mathbf{T}$, we right-shift $\mathbf{X}$ until its magnitude becomes 1 (Figure~\ref{fig:shift}). This yields the power-of-two representation $2^n$ of $\mathbf{X}$, and we compare it to $\mathbf{T}$ via:
\begin{equation}
    \left| \mathbf{X} \right| \leq \mathbf{T} \quad \Rightarrow \quad n \geq \log_2 \left( \frac{|\mathbf{X}|}{\mathbf{T}} \right)
\end{equation}

This avoids costly division by relying on fixed shift operations. The shift count can be initialized from a nonzero value for coarser estimation, effectively quantizing the threshold. This makes the technique both fast and tunable, ideal for low-power MCUs (e.g., MSP430FR5994) where multiplications and divisions are bottlenecks.

\tikzstyle{leaf}=[anchor=center, shape=circle, style={draw=blue!80, inner sep=0pt, line width=1pt, minimum size=15pt}]

\begin{figure}
    \includegraphics[width=0.48\textwidth]{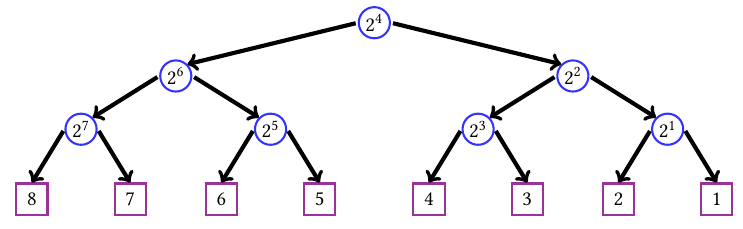}
\caption{Binary search approximation over powers of two to find the smallest n such that $2^n$ exceeds a target value. At each node, we compare and move left or right until a leaf is reached.}
\label{fig:tree}
\end{figure}

\vspace{0.5em}
\noindent\textbf{Binary Tree Search (Universal Approximation).}
Binary tree search generalizes exponent estimation through comparisons against precomputed pivot points (Figure~\ref{fig:tree}). At each level, the algorithm halves the search range, identifying the exponent of the input with respect to the pruning threshold. This supports integer, fixed-point, and reinterpreted floating-point values. Compared to bit shifting, it is slower for small numbers but more flexible—especially when input magnitudes span a wide dynamic range or exponent frequency varies.

Tree pivot points can be adapted via calibration, with frequent magnitudes occupying shallower branches. Although the lookup depth grows with input bit-width, the method is still suitable for low-precision formats. Overall, this approach is compact, portable, and memory-efficient, making it suitable for a broad class of embedded platforms.

        
        

    
\vspace{0.5em}
\noindent\textbf{Bit Masking (Floating-Point Devices).}
For platforms with floating-point units (e.g., MAX78000), \Sys applies an efficient bit-level method that leverages the IEEE 754 format. A floating-point number consists of a sign bit, an exponent field, and a mantissa:
\begin{equation}
    (-1)^S \cdot 2^{(E - E_0)} \cdot \left(1 + \frac{M}{M_{\texttt{Max}}}\right)
\end{equation}

Since the mantissa term lies in the range $\left[\frac{1}{2}, 2\right]$, we approximate the ratio of two floating-point numbers by comparing only their exponent fields:
\begin{equation}
    \left|\frac{X}{T}\right| \approx 2^{E_X - E_T}
\end{equation}

After computing the difference between the two exponents, we reapply the bias and convert the result back to a floating-point number. This gives a fast estimate of the pruning threshold without needing full division.

This yields a fast estimate of whether a connection should be pruned. The exponent is extracted via bit masking and interpreted with an offset. Because this operation requires accessing memory-resident values rather than registers, \Sys uses lookup tables for efficient implementation.

\section{Experimental Setup}
\begin{table*}[!htb]
\begin{center}
\begin{tabular}{|c|c|c|c|}
\hline
MNIST & CIFAR-10 & KWS & WiDAR\\
\hline
C: $6 \times 1 \times 5 \times 5$ & C: $6 \times 3 \times 5 \times 5$ & C: $6 \times 1 \times 5 \times 5$ & C: $32 \times 22 \times 6 \times 6$\\
P: $2 \times 2$ & P: $2 \times 2$ & P: $2 \times 2$ & C: $64 \times 32 \times 3 \times 3$ \\
C: $16 \times 6 \times 5 \times 5$ & C: $16 \times 6 \times 5 \times 5$ & C: $16 \times 6 \times 5 \times 5$ & C: $96 \times 64 \times 3 \times 3$\\
P: $2 \times 2$ & P: $2 \times 2$ & P: $2 \times 2$ & L: $1536 \times 128$\\
L: $256 \times 10$ & L: $400 \times 10$ & L: $7616 \times 12$ & L: $128 \times 6$ \\
\hline
C: Convolutional Layer & L: Linear Layer & P: Max Pool &\\
\hline
\end{tabular}
\end{center}
\caption{Model Architectures Used For Testing.}
\label{model}
\end{table*}
This section describes the experimental setup used to evaluate \Sys, including datasets, network architectures, baseline methods, and metrics for assessing MAC reduction, accuracy, runtime, and energy efficiency.

\subsection{Evaluation Platform}
We evaluate \Sys on two representative hardware platforms. First, we use the MSP430FR5994~\cite{MSP430}, an ultra-low-power microcontroller that has been widely adopted for deep inference in batteryless, energy-harvesting systems~\cite{islam2019zygarde, gobieski2019intelligence, farina2024memory}. Its ferroelectric RAM (FRAM) enables non-volatile storage with fast, low-energy access, and its limited instruction set makes multiplication significantly more expensive than branching. These properties make it an ideal candidate for evaluating MAC-skipping efficiency under extreme resource constraints. We integrate \Sys into the SONIC~\cite{gobieski2019intelligence} runtime and use TI’s EnergyTrace™ to measure real-time power consumption.

Second, we evaluate \Sys on floating-point platforms by modifying PyTorch to support runtime pruning via custom C++ backends (See Section 3.4). This reflects more general-purpose edge deployments (e.g., Raspberry Pi, Jetson Nano), where floating-point support is available but efficiency remains critical.

\subsection{Evaluation Dataset}
We evaluate \Sys across four datasets spanning core edge sensing modalities:
\begin{itemize}
    \item \textbf{MNIST}~\cite{deng2012mnist} and \textbf{CIFAR-10}~\cite{krizhevsky2009learning} for image classification, with increasing complexity.
    \item \textbf{Google Keyword Spotting (KWS)}~\cite{warden2018speech} for latency-sensitive audio inference on wake-word detection.
    \item \textbf{WiDaR}~\cite{WIDAR-dataset}, a WiFi-based indoor gesture dataset, to evaluate domain-shift robustness.
\end{itemize}

These datasets cover vision, audio, motion, and RF—common modalities in edge AI—and align with prior pruning and batteryless inference work~\cite{islam2019zygarde, farina2024memory, gobieski2019intelligence, montanari2020eperceptive}, supporting reproducibility.

Each dataset is split into training (90\%), validation (10\%), and test sets. Validation data is used only to tune pruning thresholds, and test data is reserved for final evaluation.

To assess robustness under domain shift, we use WiDaR’s protocol where training and test data come from different rooms, users, and devices. This simulates deployment drift. Specifically, we use 14 users for training and 3 for testing, with environments swapped: if Room 1 is used for training, Room 2 is used for testing, and vice versa. Room 1 is a cluttered classroom, while Room 2 is a nearly empty hallway, creating distinct signal characteristics due to environmental interference.

\subsection{Network Architecture}
We evaluate \Sys on compact convolutional neural networks tailored to the memory constraints of the MSP430FR5994 microcontroller, which provides only 256KB of FRAM. These architectures strike a balance between simplicity and representational power and are aligned with prior works such as SONIC~\cite{gobieski2019intelligence}, FreeML~\cite{farina2024memory}, and Zygarde~\cite{islam2019zygarde}. All models are initially trained using floating-point precision in PyTorch and quantized to 8-bit integers for deployment on MSP430. For floating-point evaluation, the same architecture is retained without quantization. Table~\ref{model} presents an overview of the network architecture used for each dataset.

\begin{itemize}
\item \textbf{MNIST, CIFAR-10, and KWS:}
We use 2D convolutional networks with two convolutional layers followed by max pooling and a fully connected layer. These are compact enough to run within MSP430’s fixed-point FRAM limits without model swapping.


\item \textbf{WiDar (WiFi-based Gesture Recognition):}
This model captures complex spatiotemporal CSI patterns using a LeNet-style architecture with three 2D convolutional layers followed by two linear layers~\cite{WIDAR-dataset}. Due to its larger size and floating-point requirements, it is evaluated only on desktop-class platforms and serves as a stress test for \Sys’s applicability to more complex models beyond MCU deployments.
\end{itemize}

\Sys’s pruning logic is integrated directly into the convolutional and linear layers. At inference time, each MAC operation is conditionally skipped based on a dynamic, layer-specific threshold computed per connection. This design enables fine-grained, input-aware pruning with minimal memory and compute overhead.

\begin{figure*}[!htb]
    \centering
    \includegraphics[width=0.85\textwidth]{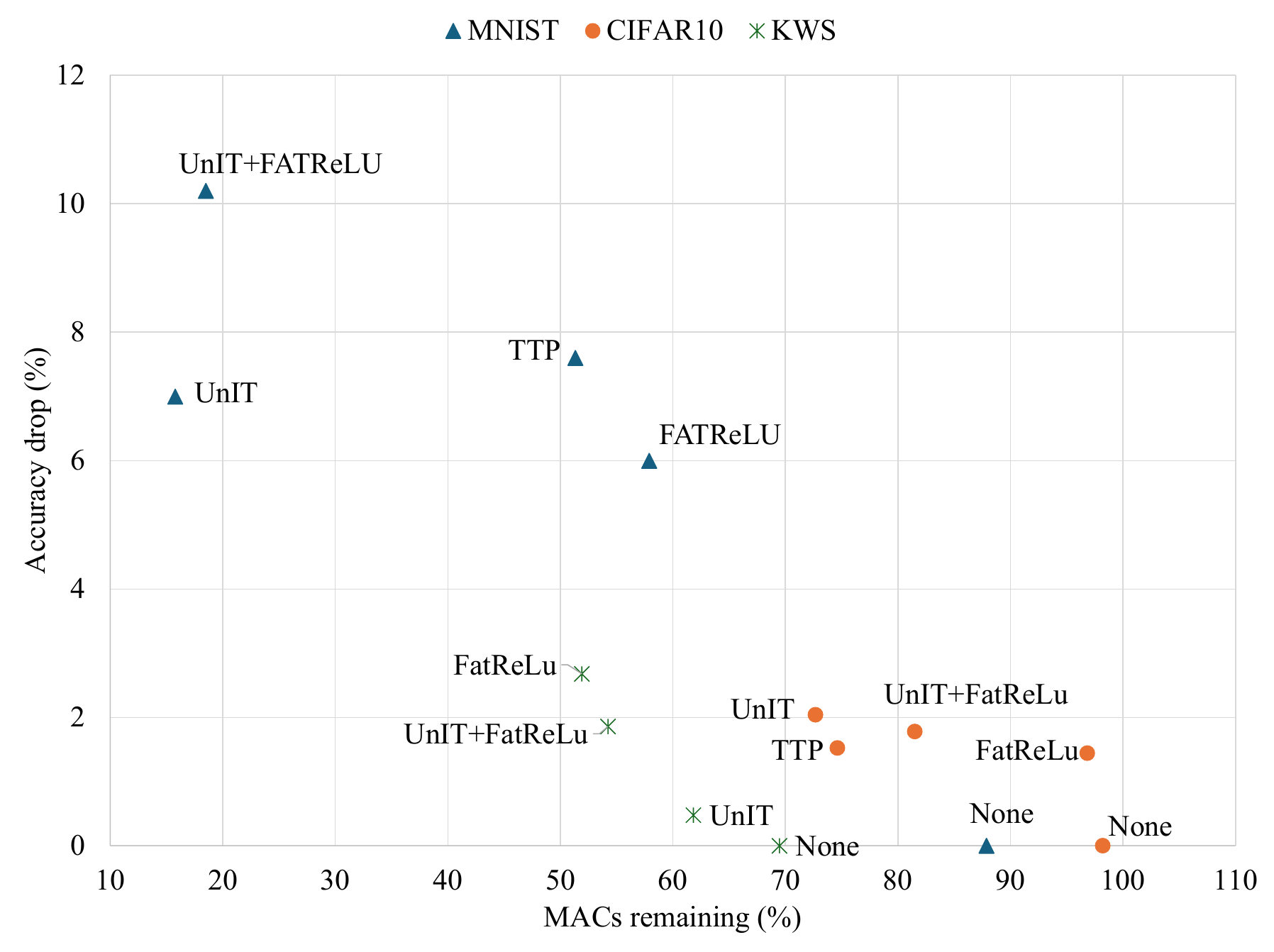}
    \caption{Accuracy drop versus remaining MAC operations across four datasets. \Sys consistently skips a high number of MACs while maintaining competitive accuracy, outperforming both train-time pruning (TTP) and inference-time pruning (FATReLU) in most cases. Combining \Sys with FATReLU further boosts sparsity without significant performance degradation. Here, ``None'' represents the non-pruned version of the model.}
    \label{fig:macsvsacc}
\end{figure*}
\subsection{Baseline}
We evaluate \Sys against the following baselines to assess its effectiveness in reducing MACs and maintaining accuracy under constrained conditions:

\begin{itemize}
    \item \textbf{Unpruned Model:} The original dense network with no pruning applied. This serves as an upper bound for accuracy and a lower bound for energy efficiency.
    \item \textbf{Training-Time Pruning:} We apply global unstructured magnitude pruning, removing weights with the smallest absolute values across the entire model. This method permanently removes parameters based on training data statistics, making it efficient but non-adaptive to runtime input variations or domain shift.
    \item \textbf{FATReLU~\cite{kurtz2020inducing}:} A state-of-the-art inference-time pruning method that modifies the standard ReLU activation function. By setting a higher threshold, FATReLU increases activation sparsity during inference, effectively skipping small activation values. It introduces minimal runtime overhead and is one of the few inference-time techniques suitable for embedded or battery-free systems.

\end{itemize}
These baselines highlight \Sys’s strengths in achieving fine-grained, adaptive pruning with low overhead, outperforming both static and runtime alternatives in efficiency and generalization.

\subsection{Evaluation metrics}
We evaluate \Sys using four different metrics.

\parlabel{Accuracy Drop} We calculate the performance drop due to the proposed method's accuracy from a baseline model. We choose the unpruned model performance as our baseline model. 
\parlabel{MACs Skipped} We calculate how many MAC operations are skipped due to the proposed method.

\parlabel{Power Consumption} We calculate the power consumption during inference with and without the overheads such as data transfer.

\parlabel{Execution Time} Similar to the power consumption, we measure the execution time for inference, including data transfer and other overhead time.

\subsection{Implementation Details}

To support inference-time pruning, we reimplement PyTorch's linear and convolutional layers in C++\footnote{https://github.com/anonymouspaper2314/UnIT-Pruner}. By customizing these layers, we gain finer control over layer operations. It enables efficient adjustment and testing of threshold values. This modification further allows faster experimentation and tuning, as C++ provides lower-level control and typically faster execution than Python alone. 

\begin{figure}
    \centering
    \subfloat[MNIST\label{mnist}]{\includegraphics[width=\linewidth]{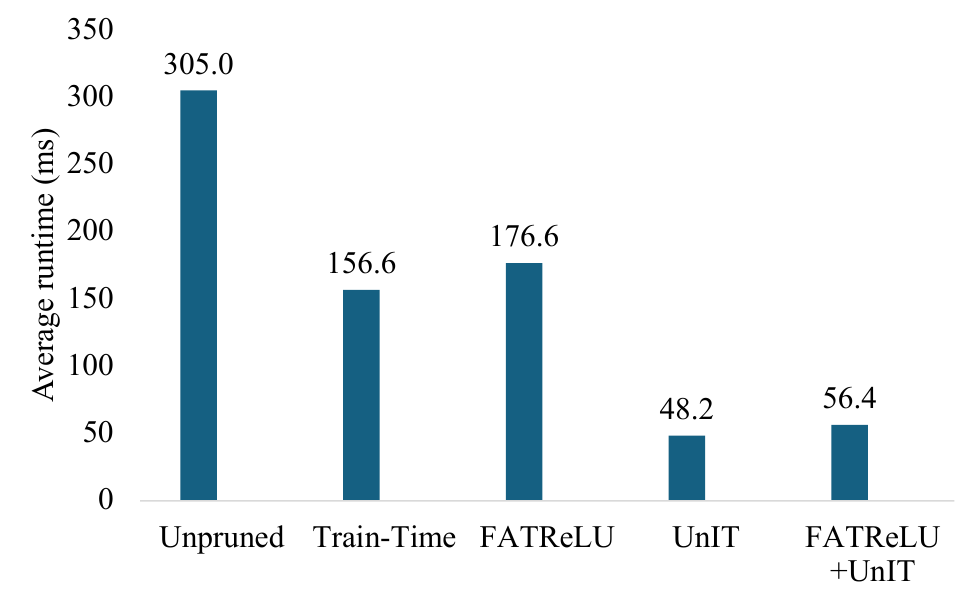}}
    
    \subfloat[CIFAR10\label{cifar10}]{\includegraphics[width=\linewidth]{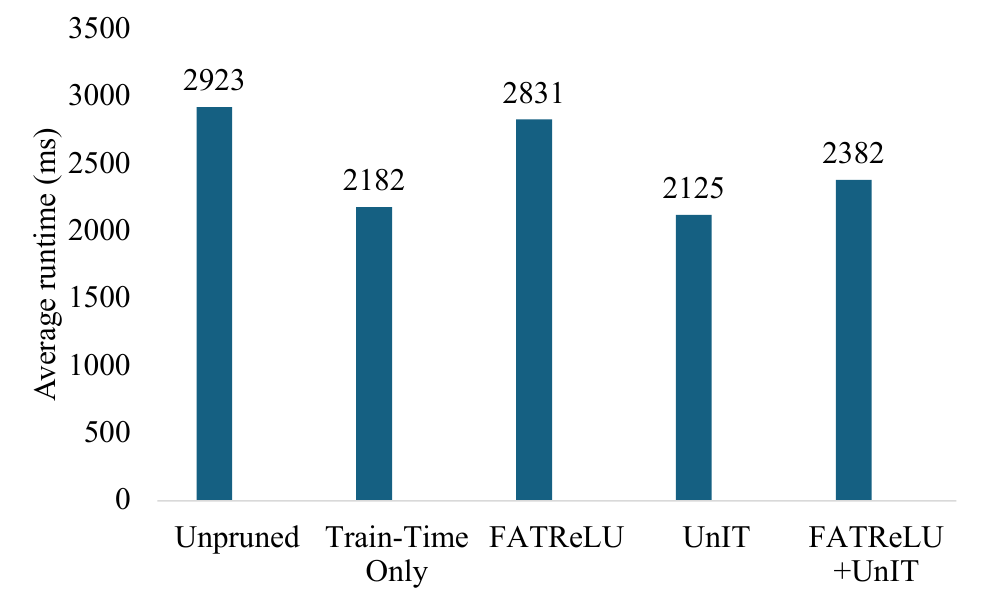}}

    \subfloat[KWS\label{kws}]{\includegraphics[width=\linewidth]{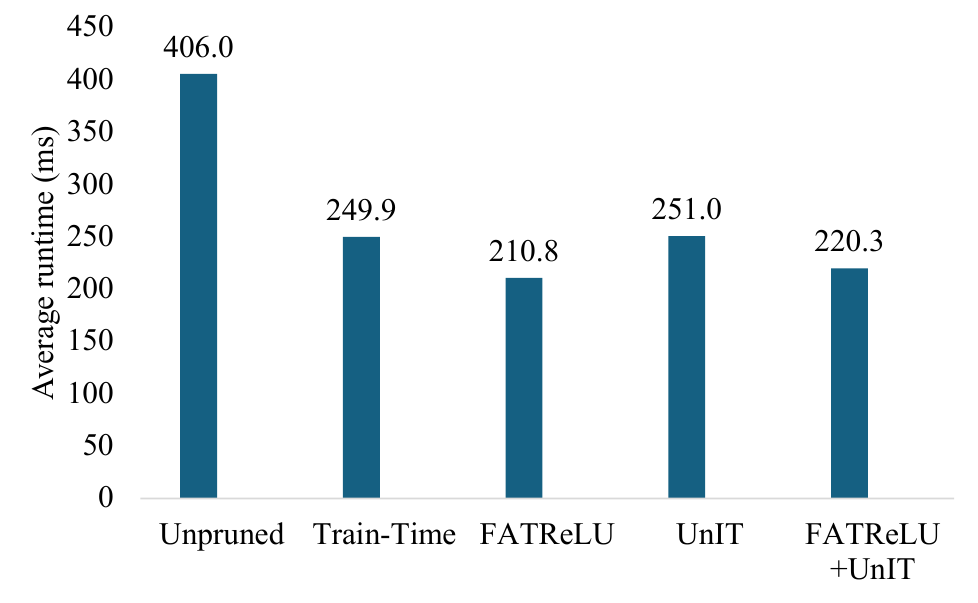}}
    \caption{Inference runtime comparison across MNIST, CIFAR10, and KWS datasets. \Sys consistently achieves lower latency than traditional train-time and inference-time pruning methods. Even when combined with FATReLU, \Sys improves runtime further, demonstrating compatibility and cumulative benefits. The runtime overhead of UnIT is 2.56~ms for MNIST, 7.52~ms for CIFAR10, and 63.52~ms for KWS, which remains small compared to the overall speedup achieved over unpruned baselines. This highlights UnIT’s efficiency for latency-sensitive MCU deployment.}
    \label{fig:time}
\end{figure}

This code is available as a Python module and will include linear, 1D convolutional, and 2D convolutional layers. In addition to the \Sys versions of the modules, we will release a debug version. This debug version hinders performance greatly but gives statistics on how many multiplications are skipped during runtime. To determine the efficiency of our division algorithms, we utilize C/C++ on its own without the support of Python.


\noindent\textbf{SONIC.}
We modify the SONIC\cite{gobieski2019intelligence} runtime to include support for \Sys. We extend their fixed-point number implementation to include our division approximation techniques. Namely, we implement the bit-shift and binary tree techniques. We modify the implementation of convolutional and linear layers to generate and utilize the thresholds as needed. We also implemented activation sparsity skipping to further demonstrate the interoperability of \Sys with other techniques.

\section{Results}
This section presents a detailed evaluation of \Sys on four diverse datasets and models deployed on microcontroller platforms. We begin by analyzing system-level efficiency, including MAC reduction, inference latency, energy consumption, and accuracy. Next, we assess \Sys’s robustness to domain shift, highlighting its adaptability in cross-context scenarios. Finally, we perform ablation studies to evaluate the performance of our hardware-efficient division approximations.

\subsection{System-Level Efficiency and Accuracy Evaluation}
\noindent\textbf{Computation Reduction and Accuracy Retention.}
Figure~\ref{fig:macsvsacc} shows the model performance with different pruning methods for the four datasets. \Sys skips 27.30 - 84.21\% MAC operation with only 0.48 - 7\% accuracy drop. Our highest accuracy drop (7\%) is with MNIST, where we skipped 84.21\% of the multiplications. For the KWS dataset, the maximum accuracy drop is only 0.48\% with skipping 38.8\% MAC operation. Overall, \Sys skips many unnecessary multiplications without a significant accuracy drop, which is crucial for low-power MCUs.

\begin{figure}
    \centering
    \subfloat[MNIST\label{mnist_e}]{\includegraphics[width=\linewidth]{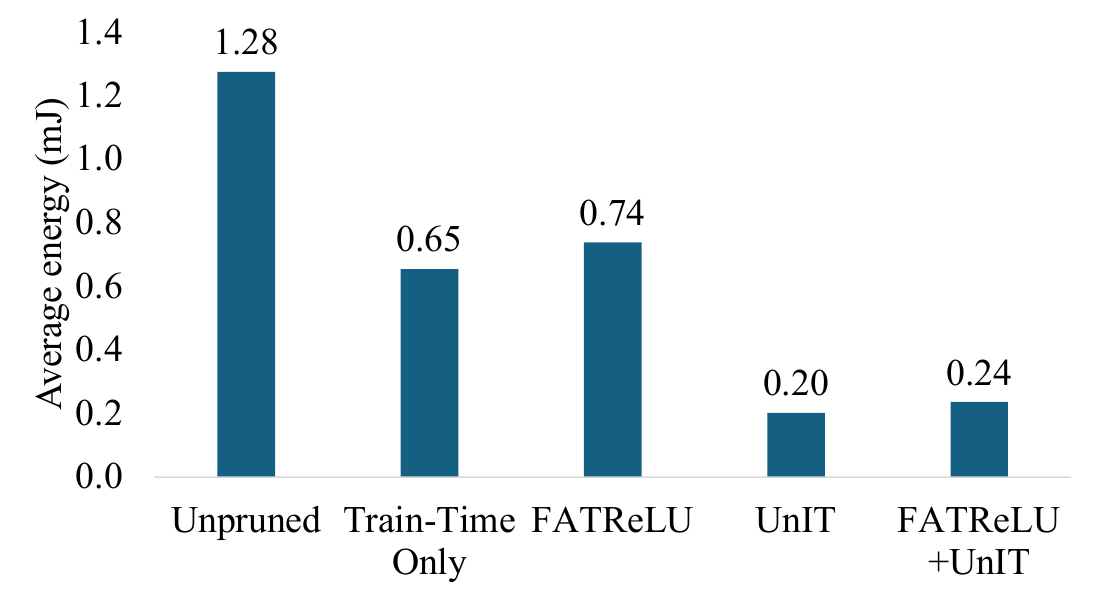}}

    \subfloat[CIFAR10\label{cifar10_e}]{\includegraphics[width=\linewidth]{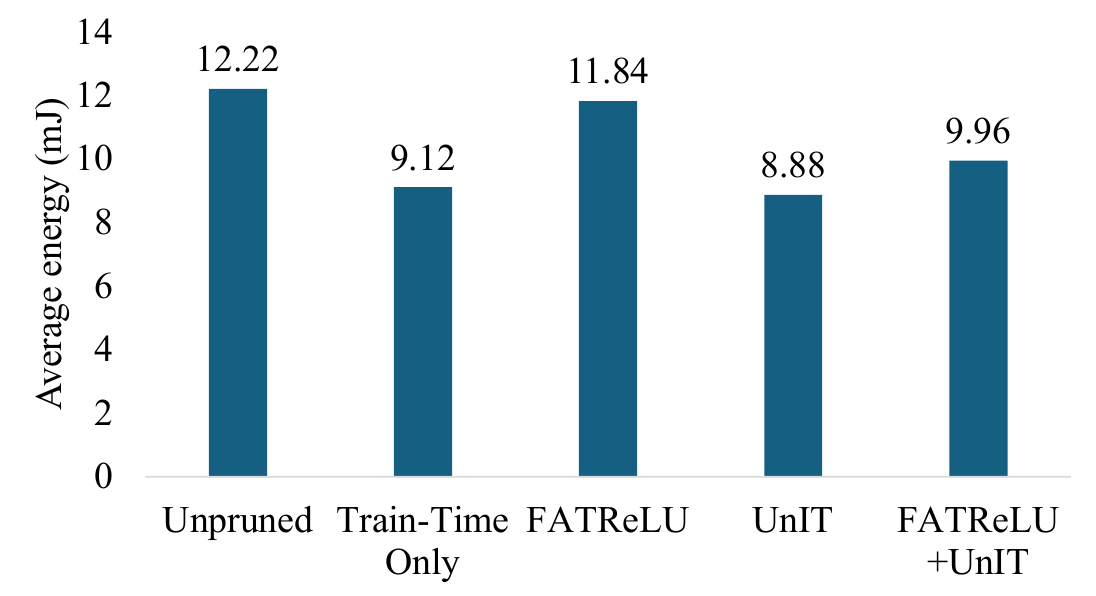}}

    \subfloat[KWS\label{kws_e}]{\includegraphics[width=\linewidth]{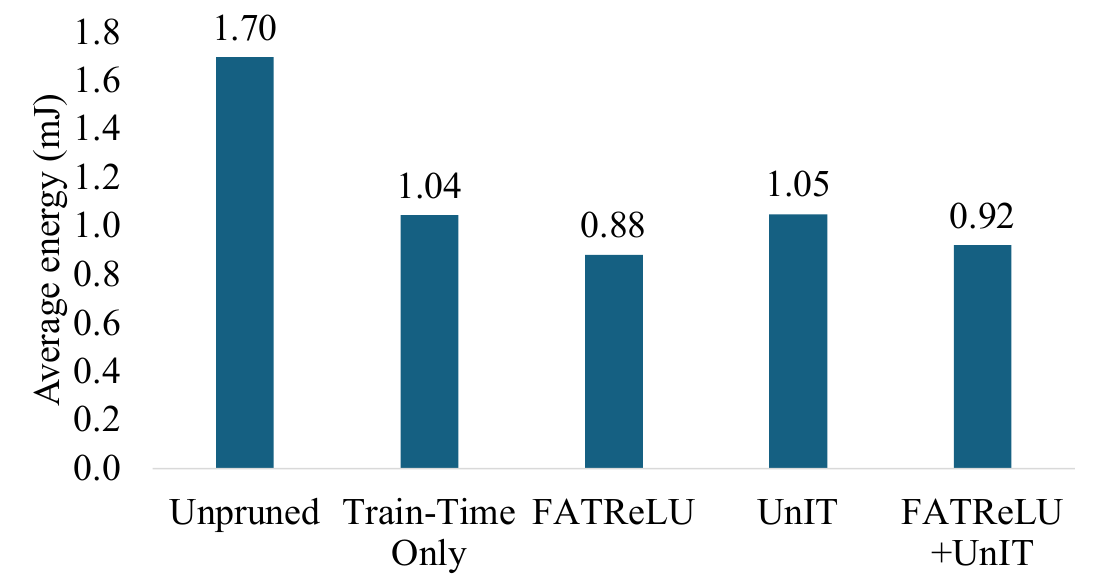}}
    \caption{Average energy consumption during inference across MNIST, CIFAR10, and KWS datasets. UnIT consistently achieves lower energy usage than both train-time pruning and FATReLU due to its input-aware MAC skipping. On MNIST, UnIT reduces energy from 1.28 mJ to 0.20 mJ, yielding an 84\% reduction. On CIFAR10, it reduces energy by 27\% compared to the unpruned model, and it performs competitively on KWS. When combined with FATReLU, UnIT further improves energy efficiency in some settings, highlighting its compatibility with other pruning strategies and its potential for cumulative benefits.}
    \label{fig:energy}
\end{figure}

Moreover, the proposed method has 5.85--42.2\% less MAC operation compared to train-time pruning with a -0.06--0.65\% more accuracy drop. Compared to FATReLU, our model has 7.69\% fewer MAC operations with 0.6\% less accuracy in CIFAR-10. On the other three datasets, \Sys achieves 0.63--2.2\% more accuracy with 20.06-50\% less MAC operation. Thus, our method can be more efficient with fewer MAC operations without sacrificing the performance.

\Sys not only achieves comparable performance when skipping the vast majority of MAC operations, but is also a more efficient pruning method for low-powered MCUs than FATReLU. Figure \ref{fig:macsvsacc} shows that our method not only outperforms FATReLU but integrating FATReLU with \Sys does not give any advantage, making our method the most effective in these tests.



\noindent\textbf{Latency Performance.}
Figure \ref{fig:time} shows the time needed for inference along with the data moving time for different methods of model compression on multiple datasets. Figure~\ref{fig:time} shows that \Sys drastically reduces inference time by 27.30--84.19\% compared to train-time pruning and 24.9--72.7\% faster than FATReLU on CIFAR10 and MNIST datasets, achieving times of 7.5--151 and 3.8--125 seconds, respectively. We also observe that \Sys has a lower time overhead than FATReLU. As expected for a battery-free system, most of the time is spent moving data and performing other computations. 






\noindent\textbf{Energy Efficiency Under Pruning.}
In Figure~\ref{fig:energy}, UnIT consumes only 0.20--8.8mJ per inference on the MSP430, whereas FATReLU needs 0.74--11.84mJ, and train-time pruning needs 0.65--12.22mJ. Even with the energy for data transfer, overhead, and other computational tasks, our algorithm saves the most energy when compared to other methods, making it suitable for low-powered and battery-free systems.

\begin{table*}[]
    \centering
    \caption{Effectiveness of \Sys under contextual shifts, evaluated using the WIDAR dataset collected from two distinct environments (Room 1 and Room 2). Despite changes in data distribution between rooms, \Sys consistently achieves over 60\% MAC reduction with minimal F1 score drop. Combining \Sys with train-time pruning yields the highest MAC savings with negligible additional performance degradation.}
    \resizebox{\linewidth}{!}{
    \begin{tabular}{|c|c|c|c|c|c|c|c|c|}
    \hline
    Training Context $\rightarrow$ & \multicolumn{4}{|c|}{Room 1}& \multicolumn{4}{|c|}{Room 2}\\
    \hline
         Testing Context $\rightarrow$& \multicolumn{2}{|c|}{Room 1}& \multicolumn{2}{|c|}{Room 2}& \multicolumn{2}{|c|}{Room 1}& \multicolumn{2}{|c|}{Room 2} \\
         \hline
         Mechanism $\downarrow$ /Metric $\rightarrow$& F1 score & MAC Skipped \% & F1 score & MAC Skipped \% & F1 score & MAC Skipped \% & F1 score & MAC Skipped \% \\
         \hline
         Unpruned & 0.4568 & 16.50  & 0.6915& 16.37 &0.4353& 17.19 & 0.5837 &17.07\\
         Train-time Only & 0.4412& 55.14& 0.7061& 55.05 & 0.4220& 53.61 &0.6049&53.53\\
         UnIT &  0.4650 & 63.31 & 0.7016 &61.86 & 0.4282& 64.31& 0.5773 & 62.65\\
         Train-time Only + UnIT & 0.4419& 71.50 & 0.7079 & 70.08 &0.4193 & 72.04& 0.5961& 70.60\\
         \hline
    \end{tabular}}
    
    \label{tab:widar}
\end{table*}

\subsection{Cross-Context Robustness and Adaptivity Evaluation}
This section evaluates the robustness of \Sys under domain shift by analyzing its ability to reduce MAC operations while preserving model performance, measured using the F1 score. Table~\ref{tab:widar} presents results for models trained in one environment (Room 1 or Room 2) and tested on disjoint data from the alternate setting. Because \Sys makes pruning decisions dynamically based on each input, it allows models to generalize more effectively across deployment conditions that differ from the training distribution.

Across all setups, the F1 scores remain within $\pm 1\%$ of the original unpruned model, confirming that \Sys maintains accuracy under both same-context and cross-context evaluation. While train-time pruning alone reduces MAC operations substantially, \Sys consistently achieves greater MAC reduction without sacrificing performance. When applied in conjunction with train-time pruning, it enables additional savings, demonstrating that \Sys can complement other pruning strategies for further efficiency gains. Importantly, the effectiveness of \Sys remains stable across distinct data distributions, indicating that its adaptivity generalizes well under deployment drift and real-world variability.

\subsection{Efficiency of Division Approximation}

\begin{figure}
    \centering
    \includegraphics[width=0.43\textwidth]{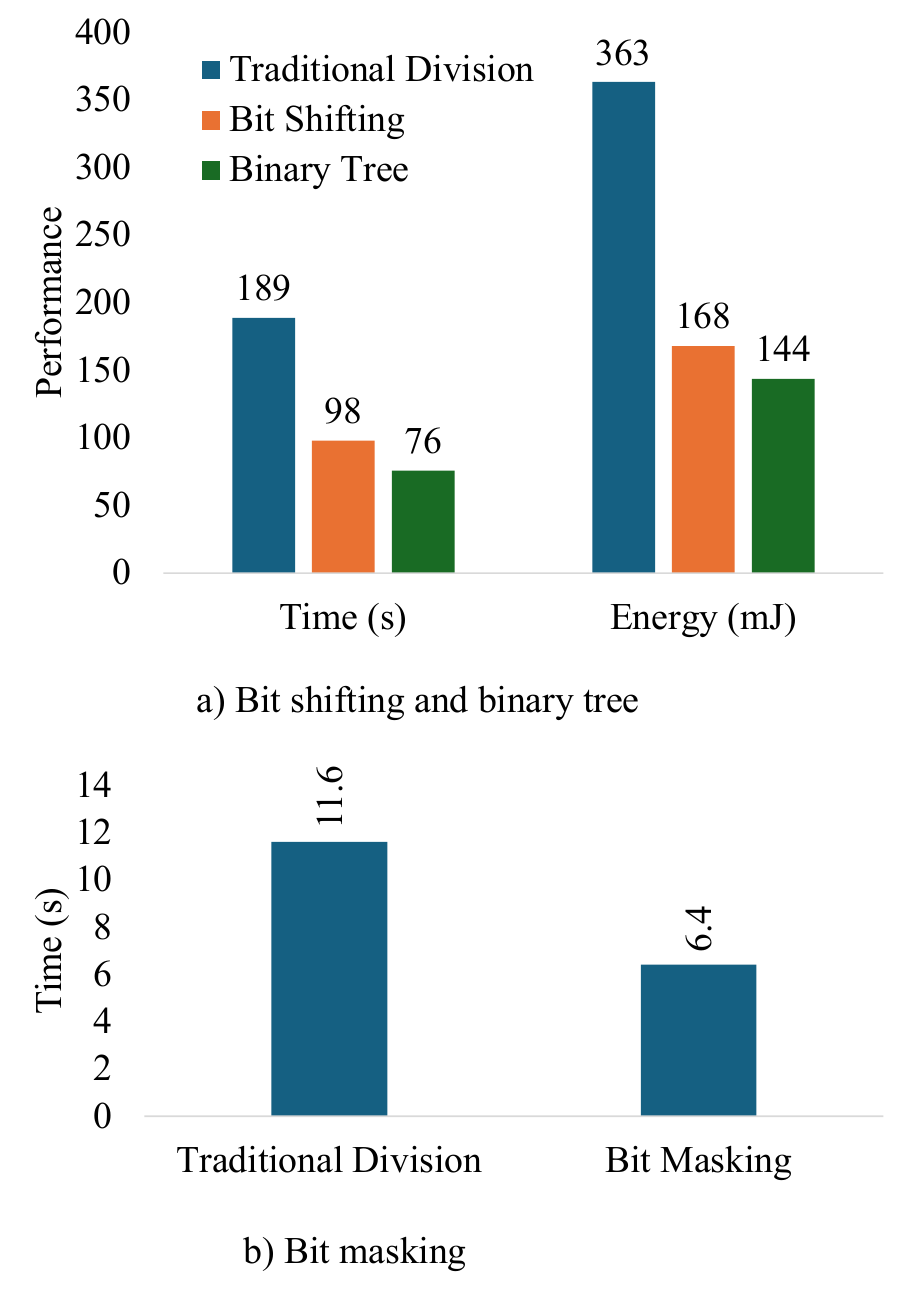}
    \vspace{-1em}
    \caption{Approximate division methods using bit shifting, binary tree search, and bit masking significantly reduce runtime and energy consumption compared to traditional division, offering faster and more efficient computation.}
    \label{fig:division}
\end{figure}

We compare the performance of our fast division approximation techniques with that of traditional divisions. We perform bit shifting and binary tree searches on the MSP430 as they are suitable for integer and fixed point systems. Although we cannot compare the performance of the bit masking method with the traditional division on the MSP430 due to the unavailability of a floating point system, we perform bit masking on an Intel(R) Core(TM) i7-9750H CPU @ 2.60GHz, 2592 Mhz, 6 Core(s), 12 Logical Processor(s) for 10 billion iterations. Figure~\ref{fig:division} shows the performance of our fast approximate division techniques compared to the traditional division. Bit shifting and binary tree search both achieve 50-59.8\% lower execution time and 53.7 - 60.3\% lower energy consumption when compared to the conventional division method on the MSP430. Similarly, bit masking completes 44.8\% faster than the traditional method.







\section{Related Work}
\parlabel{DNN in Low-powered MCUs}
DNN in low-powered MCUs requires additional processing to make the model memory and computationally efficient. Researchers have deployed train-time pruning \cite{liu2018rethinking}, special pruning \cite{han2015deep}, and quantization \cite{wu2016quantized} to make the model smaller. Tensor decomposition and early-exit strategies have been popular in reducing inference time for on-device computation ~\cite{teerapittayanon2016branchynet, islam2019zygarde, montanari2020eperceptive, farina2024memory}. Neural architecture search (NAS) is used to find the most compressed model to run on low-resourced systems ~\cite{gobieski2019intelligence, lyu2021resource}. Some systems utilize Knowledge distillation to make a smaller model from a larger model \cite{bharadhwaj2022detecting, qi2022fedbkd}. Researchers also utilized a combination of quantization or pruning with knowledge distillation for a more efficient model \cite{qu2020quantization}.

\parlabel{Structured Pruning}
Structured pruning targets the removal of entire structures, such as neurons, channels, or even layers. This approach simplifies model architecture in a way that aligns well with hardware, as it results in smaller matrices or tensor dimensions. Notable structured pruning methods include \textit{channel pruning}~\cite{li2017pruning}, which removes channels based on a criterion like weight magnitude or contribution to overall model accuracy. \textit{Filter pruning}~\cite{luo2017thinet} is another structured technique that removes filters based on the importance of their activations, reducing convolutional layer sizes in CNNs and enhancing deployment speed on hardware that benefits from reduced channel dimensions. More recent structured pruning techniques also involve \textit{group sparsity regularization}~\cite{wen2016learning}, where specific regularizers are applied to groups of weights to force sparsity within predefined structures, yielding efficient models without major loss in accuracy. Structured pruning is widely favored for deployment on accelerators, as it produces compact models without the need for additional computation to handle irregular sparsity. 

\parlabel{Unstructured Pruning}
Unstructured pruning, by contrast, removes individual weights based on specific criteria, such as magnitude. The \textit{magnitude-based pruning} approach~\cite{han2015learning} is one of the most widely used methods in this category, where weights with the smallest absolute values are removed iteratively, yielding a sparse weight matrix. This technique, while effective in significantly reducing model parameters, results in irregular sparsity patterns that are less efficient to compute on standard hardware due to the need for specialized libraries and frameworks to leverage the sparsity. Techniques such as \textit{variational dropout}~\cite{molchanov2017variational} use Bayesian methods to apply pruning during training, automatically determining which weights can be removed by learning a probability distribution over them. Although unstructured pruning can yield highly sparse networks, the irregularity of the resulting weights often limits practical speedup benefits during deployment. \citet{no-time-reduction} supported this claim by showing higher efficiency obtained using filter pruning (a form of structure pruning) than weight pruning (unstructured pruning) without the use of specialized hardware or libraries.

\parlabel{Global vs. Layer-wise Thresholding}
Global thresholding applies a single pruning threshold across all layers of a model, while layer-wise thresholding assigns separate thresholds per layer. Prior work has shown that weight distributions vary significantly across layers, making global thresholds suboptimal~\cite{learned-threshold}. \citet{10.1007/978-3-030-63823-8_58} argue for layer-wise thresholding as a more adaptive strategy, and recent methods~\cite{zhao2023automaticattentionpruningimproving} have adopted this approach to improve pruning effectiveness. Our design similarly embraces layer-wise thresholding to better accommodate layer-specific characteristics.

\parlabel{Training-Time Pruning}
Training-time pruning, also known as \textit{dynamic pruning}, integrates pruning operations within the training process. This category includes \textit{Gradual Magnitude Pruning} (GMP)~\cite{zhu2017to}, where weights are pruned iteratively throughout the training process, allowing the model to adapt and learn with progressively fewer parameters. Similarly, the \textit{Lottery Ticket Hypothesis}~\cite{frankle2018lottery} suggests identifying subnetworks with randomly initialized parameters that can be retrained to achieve near-original accuracy, leading to a more efficient training process. Training-time pruning can also be applied through \textit{regularization-based techniques}~\cite{liu2017rethinking}, where sparsity-inducing norms (e.g., $L_1$ norm) are added to the loss function, encouraging the model to develop sparse weight patterns inherently. By pruning during training, these approaches enable the model to adapt to lower parameter counts while preserving accuracy, though they may add complexity to the training process.

\parlabel{Post-Training Pruning}
\textit{Post-training pruning} is applied to pre-trained models allowing practitioners to achieve model size reductions without altering the original training pipeline. In this category, \textit{post-training quantization and pruning}~\cite{hubara2016binarized} is a widely used method, where redundant parameters are pruned after training and then quantized to reduce memory footprint, resulting in efficient deployment. \textit{Layer-wise pruning}~\cite{liu2019sparse} is another post-training approach that uses sensitivity analysis on each layer of a trained network to prune layers with minimal impact on model performance. Techniques such as \textit{fine-tuning after pruning}~\cite{han2015learning} are commonly employed to recover any accuracy loss following pruning, where the model is slightly retrained after pruning to regain accuracy lost due to the reduction in parameters.

\parlabel{Inference-Time Pruning}
Inference-time pruning refers to techniques that are done \emph{during} inference. Techniques like FATReLU~\cite{kurtz2020inducing}, also known as Truncated Rectified~\cite{konda2015zerobiasautoencodersbenefitscoadapting}, which induce more sparsity only during runtime are examples of inference-time neuron pruning. There has also been work involving inference-time channel pruning~\cite{NIPS2017_a51fb975} where a simpler model was trained in tandem with the main model to signal when to skip certain filters. These methods are all structured, making \Sys the first of its kind as an unstructured inference-time pruning system.

\section{Discussion and  Limitations}
This section discusses the shortcomings and future potential of \Sys. 

\subsection{Trade-Off Between Adaptability and Network Size}
One notable advantage of \Sys is its ability to retain activations and weights for potential future use, offering greater adaptability compared to traditional train-time and structured pruning methods. This flexibility is especially beneficial in environments where computational and energy resources fluctuate. However, preserving these weights limits the ability to reduce the network size further. This presents a challenge, particularly given the algorithm's primary focus on resource-constrained hardware, where minimizing network size is crucial for optimal performance.


\subsection{\Sys on Highly Parallelized Hardware}

GPUs and other highly parallelized hardware may struggle to leverage \Sys fully. In such systems, the potential savings are diminished because all cores must wait for the slowest one, causing the worst-performing core to dictate the speed of the operation. This blocking behavior reduces the effectiveness of the algorithm on these devices. While the results presented here suggest that the frequency of skipped operations could be high enough for certain systems to benefit, it is evident that as the number of cores increases, the overall savings will decrease. Future experiments will be conducted to verify this hypothesis.

\subsection{Portability Limitations Due to Hardware-Specific Division Techniques}
Proposed fast division approximation techniques significantly reduce computational overhead, making the algorithm suitable for many embedded systems. However, the need for different algorithms depending on the hardware used means that the portability of \Sys runtimes would be limited. This hardware dependency may restrict the algorithm's versatility across diverse platforms and complicate its deployment in varied environments.

\subsection{Impact of Model Size}
Since the algorithm was designed with low-power systems in mind, the testing so far has been focused on smaller models to ensure efficient resource utilization. However, the performance and benefits of \Sys on larger, more complex models remain untested. In future work, we plan to evaluate \Sys with larger models to better understand its scalability, performance under higher computational loads, and ability to maintain efficiency across a broader range of applications. This will provide a more comprehensive assessment of the algorithm's full potential and help identify any additional challenges that may arise with increased model complexity.

\subsection{Expand to Diverse DNN Layers}
Thus far, we have only tested \Sys with convolutional and linear layers, primarily due to their popularity and efficiency, especially in resource-constrained applications. Future work will include testing a broader range of layer types, such as recurrent and attention layers, to assess \ Sys's adaptability and performance across different architectures. This will help evaluate its general applicability to a broader set of neural network models.

\subsection{Testing on Diverse Edge Devices}
While the MSP430 is a good starting point, we aim to test \Sys on a wider range of devices, particularly edge computing devices with unique properties. We believe the flexibility of the techniques described here will be instrumental in porting the algorithm to various embedded systems, enabling its use in different hardware environments. This will help assess the adaptability and performance of \Sys across a broader spectrum of edge devices, contributing to its potential for more widespread deployment

\subsection{Data Transfer Overhead}
Although our method reduces inference time significantly, the energy and time spent moving memory on the device currently outweigh the time spent on computations, resulting in high total inference time. We plan to explore alternative, application-specific memory management systems tailored for ultra-low-power devices. Future work could investigate more efficient storage and retrieval mechanisms that better support dynamic pruning while maintaining the benefits of adaptability. Additionally, we aim to develop techniques to minimize memory transfer overhead, further optimizing the overall performance of \Sys on resource-constrained hardware.

\subsection{Imprecise BLAS}
Fundamentally, the techniques outlined here for quickly generating thresholds and applying them induce and take advantage of sparsity in matrix multiplications. While these threshold generation techniques may not be as valuable in machine learning contexts for systems with ample storage, they can still be useful in scenarios like Basic Linear Algebra Subprograms (BLAS), where the two matrix values are entirely unknown. In such cases, these algorithms may offer a form of dynamic thresholding, helping to improve efficiency by selectively skipping redundant operations. This approach highlights the potential applicability of these methods beyond traditional machine learning tasks, particularly in resource-constrained environments.

\section{Conclusion}
We present \Sys, a lightweight, unstructured inference-time pruning method tailored for deployment on low-power microcontrollers (MCUs). By dynamically skipping individual MAC operations based on input-aware thresholds, \Sys eliminates unnecessary computation without requiring retraining or altering the model architecture. Its reuse-aware thresholding and hardware-efficient division approximations enable fine-grained pruning with minimal overhead. Evaluations across four diverse datasets demonstrate that \Sys consistently reduces MAC operations, energy consumption, and latency—even under domain shift—making it a practical and effective solution for efficient inference on resource-constrained MCU platforms.
Looking ahead, we plan to extend \Sys’s capabilities to larger models and investigate its integration with other pruning strategies to further optimize performance across a broader range of applications.



\bibliographystyle{ACM-Reference-Format}
\bibliography{main}


\begin{thebibliography}{46}


\ifx \showCODEN    \undefined \def \showCODEN     #1{\unskip}     \fi
\ifx \showDOI      \undefined \def \showDOI       #1{#1}\fi
\ifx \showISBNx    \undefined \def \showISBNx     #1{\unskip}     \fi
\ifx \showISBNxiii \undefined \def \showISBNxiii  #1{\unskip}     \fi
\ifx \showISSN     \undefined \def \showISSN      #1{\unskip}     \fi
\ifx \showLCCN     \undefined \def \showLCCN      #1{\unskip}     \fi
\ifx \shownote     \undefined \def \shownote      #1{#1}          \fi
\ifx \showarticletitle \undefined \def \showarticletitle #1{#1}   \fi
\ifx \showURL      \undefined \def \showURL       {\relax}        \fi
\providecommand\bibfield[2]{#2}
\providecommand\bibinfo[2]{#2}
\providecommand\natexlab[1]{#1}
\providecommand\showeprint[2][]{arXiv:#2}

\bibitem[MSP(2018)]%
        {MSP430Mul}
 \bibinfo{year}{2018}\natexlab{}.
\newblock \bibinfo{title}{Efficient Multiplication and Division Using MSP430™ MCUs}.
\newblock
\newblock
\urldef\tempurl%
\url{https://www.ti.com/lit/an/slaa329a/slaa329a.pdf}
\showURL{%
\tempurl}


\bibitem[MSP(2024)]%
        {MSP430}
 \bibinfo{year}{2024}\natexlab{}.
\newblock \bibinfo{title}{MSP430 User Guide}.
\newblock
\newblock
\urldef\tempurl%
\url{https://www.ti.com/sc/docs/products/micro/msp430/userguid/as_5.pdf}
\showURL{%
\tempurl}


\bibitem[Anwar et~al\mbox{.}(2017)]%
        {anwar2017structured}
\bibfield{author}{\bibinfo{person}{Sajid Anwar}, \bibinfo{person}{Kyuyeon Hwang}, {and} \bibinfo{person}{Wonyong Sung}.} \bibinfo{year}{2017}\natexlab{}.
\newblock \showarticletitle{Structured pruning of deep convolutional neural networks}.
\newblock \bibinfo{journal}{\emph{ACM Journal on Emerging Technologies in Computing Systems (JETC)}} \bibinfo{volume}{13}, \bibinfo{number}{3} (\bibinfo{year}{2017}), \bibinfo{pages}{1--18}.
\newblock


\bibitem[Azarian et~al\mbox{.}(2020)]%
        {learned-threshold}
\bibfield{author}{\bibinfo{person}{Kambiz Azarian}, \bibinfo{person}{Yash Bhalgat}, \bibinfo{person}{Jinwon Lee}, {and} \bibinfo{person}{Tijmen Blankevoort}.} \bibinfo{year}{2020}\natexlab{}.
\newblock \bibinfo{title}{Learned Threshold Pruning}.
\newblock
\newblock
\urldef\tempurl%
\url{https://doi.org/10.48550/arXiv.2003.00075}
\showDOI{\tempurl}


\bibitem[Bakar et~al\mbox{.}(2022)]%
        {bakar2022protean}
\bibfield{author}{\bibinfo{person}{Abu Bakar}, \bibinfo{person}{Rishabh Goel}, \bibinfo{person}{Jasper De~Winkel}, \bibinfo{person}{Jason Huang}, \bibinfo{person}{Saad Ahmed}, \bibinfo{person}{Bashima Islam}, \bibinfo{person}{Przemys{\l}aw Pawe{\l}czak}, \bibinfo{person}{Kas{\i}m~Sinan Y{\i}ld{\i}r{\i}m}, {and} \bibinfo{person}{Josiah Hester}.} \bibinfo{year}{2022}\natexlab{}.
\newblock \showarticletitle{Protean: An energy-efficient and heterogeneous platform for adaptive and hardware-accelerated battery-free computing}. In \bibinfo{booktitle}{\emph{Proceedings of the 20th ACM Conference on Embedded Networked Sensor Systems}}. \bibinfo{pages}{207--221}.
\newblock


\bibitem[Bharadhwaj et~al\mbox{.}(2022)]%
        {bharadhwaj2022detecting}
\bibfield{author}{\bibinfo{person}{Manoj Bharadhwaj}, \bibinfo{person}{Gitakrishnan Ramadurai}, {and} \bibinfo{person}{Balaraman Ravindran}.} \bibinfo{year}{2022}\natexlab{}.
\newblock \showarticletitle{Detecting vehicles on the edge: Knowledge distillation to improve performance in heterogeneous road traffic}. In \bibinfo{booktitle}{\emph{Proceedings of the IEEE/CVF Conference on Computer Vision and Pattern Recognition}}. \bibinfo{pages}{3192--3198}.
\newblock


\bibitem[Biswas et~al\mbox{.}(2025)]%
        {biswas2025quads}
\bibfield{author}{\bibinfo{person}{Subrata Biswas}, \bibinfo{person}{Mohammad Nur~Hossain Khan}, {and} \bibinfo{person}{Bashima Islam}.} \bibinfo{year}{2025}\natexlab{}.
\newblock \showarticletitle{QUADS: QUAntized Distillation Framework for Efficient Speech Language Understanding}.
\newblock \bibinfo{journal}{\emph{arXiv preprint arXiv:2505.14723}} (\bibinfo{year}{2025}).
\newblock


\bibitem[Cheng et~al\mbox{.}(2017)]%
        {cheng2017survey}
\bibfield{author}{\bibinfo{person}{Yu Cheng}, \bibinfo{person}{Duo Wang}, \bibinfo{person}{Pan Zhou}, {and} \bibinfo{person}{Tao Zhang}.} \bibinfo{year}{2017}\natexlab{}.
\newblock \showarticletitle{A survey of model compression and acceleration for deep neural networks}.
\newblock \bibinfo{journal}{\emph{arXiv preprint arXiv:1710.09282}} (\bibinfo{year}{2017}).
\newblock


\bibitem[Deng(2012)]%
        {deng2012mnist}
\bibfield{author}{\bibinfo{person}{Li Deng}.} \bibinfo{year}{2012}\natexlab{}.
\newblock \showarticletitle{The mnist database of handwritten digit images for machine learning research}.
\newblock \bibinfo{journal}{\emph{IEEE Signal Processing Magazine}} \bibinfo{volume}{29}, \bibinfo{number}{6} (\bibinfo{year}{2012}), \bibinfo{pages}{141--142}.
\newblock


\bibitem[Diab and Rodriguez-Villegas(2022)]%
        {diab2022embedded}
\bibfield{author}{\bibinfo{person}{Maha~S Diab} {and} \bibinfo{person}{Esther Rodriguez-Villegas}.} \bibinfo{year}{2022}\natexlab{}.
\newblock \showarticletitle{Embedded machine learning using microcontrollers in wearable and ambulatory systems for health and care applications: A review}.
\newblock \bibinfo{journal}{\emph{IEEE Access}}  \bibinfo{volume}{10} (\bibinfo{year}{2022}), \bibinfo{pages}{98450--98474}.
\newblock


\bibitem[Farina et~al\mbox{.}(2024)]%
        {farina2024memory}
\bibfield{author}{\bibinfo{person}{Pietro Farina}, \bibinfo{person}{Subrata Biswas}, \bibinfo{person}{Eren Y{\i}ld{\i}z}, \bibinfo{person}{Khakim Akhunov}, \bibinfo{person}{Saad Ahmed}, \bibinfo{person}{Bashima Islam}, {and} \bibinfo{person}{Kas{\i}m~Sinan Y{\i}ld{\i}r{\i}m}.} \bibinfo{year}{2024}\natexlab{}.
\newblock \showarticletitle{Memory-efficient Energy-adaptive Inference of Pre-Trained Models on Batteryless Embedded Systems}.
\newblock \bibinfo{journal}{\emph{arXiv preprint arXiv:2405.10426}} (\bibinfo{year}{2024}).
\newblock


\bibitem[Frankle and Carbin(2018)]%
        {frankle2018lottery}
\bibfield{author}{\bibinfo{person}{Jonathan Frankle} {and} \bibinfo{person}{Michael Carbin}.} \bibinfo{year}{2018}\natexlab{}.
\newblock \showarticletitle{The lottery ticket hypothesis: Finding sparse, trainable neural networks}. In \bibinfo{booktitle}{\emph{International Conference on Learning Representations}}.
\newblock


\bibitem[Gobieski et~al\mbox{.}(2019)]%
        {gobieski2019intelligence}
\bibfield{author}{\bibinfo{person}{Graham Gobieski}, \bibinfo{person}{Brandon Lucia}, {and} \bibinfo{person}{Nathan Beckmann}.} \bibinfo{year}{2019}\natexlab{}.
\newblock \showarticletitle{Intelligence beyond the edge: Inference on intermittent embedded systems}. In \bibinfo{booktitle}{\emph{Proceedings of the Twenty-Fourth International Conference on Architectural Support for Programming Languages and Operating Systems}}. \bibinfo{pages}{199--213}.
\newblock


\bibitem[Gorlatova et~al\mbox{.}(2010)]%
        {gorlatova2010energy}
\bibfield{author}{\bibinfo{person}{Maria Gorlatova}, \bibinfo{person}{Peter Kinget}, \bibinfo{person}{Ioannis Kymissis}, \bibinfo{person}{Dan Rubenstein}, \bibinfo{person}{Xiaodong Wang}, {and} \bibinfo{person}{Gil Zussman}.} \bibinfo{year}{2010}\natexlab{}.
\newblock \showarticletitle{Energy harvesting active networked tags (EnHANTs) for ubiquitous object networking}.
\newblock \bibinfo{journal}{\emph{IEEE Wireless Communications}} \bibinfo{volume}{17}, \bibinfo{number}{6} (\bibinfo{year}{2010}), \bibinfo{pages}{18--25}.
\newblock


\bibitem[Han et~al\mbox{.}(2015a)]%
        {han2015deep}
\bibfield{author}{\bibinfo{person}{Song Han}, \bibinfo{person}{Huizi Mao}, {and} \bibinfo{person}{William~J Dally}.} \bibinfo{year}{2015}\natexlab{a}.
\newblock \showarticletitle{Deep compression: Compressing deep neural networks with pruning, trained quantization and huffman coding}.
\newblock \bibinfo{journal}{\emph{arXiv preprint arXiv:1510.00149}} (\bibinfo{year}{2015}).
\newblock


\bibitem[Han et~al\mbox{.}(2015b)]%
        {han2015learning}
\bibfield{author}{\bibinfo{person}{Song Han}, \bibinfo{person}{Jeff Pool}, \bibinfo{person}{John Tran}, {and} \bibinfo{person}{William Dally}.} \bibinfo{year}{2015}\natexlab{b}.
\newblock \showarticletitle{Learning both weights and connections for efficient neural network}.
\newblock \bibinfo{journal}{\emph{Advances in neural information processing systems}}  \bibinfo{volume}{28} (\bibinfo{year}{2015}).
\newblock


\bibitem[Hubara et~al\mbox{.}(2016)]%
        {hubara2016binarized}
\bibfield{author}{\bibinfo{person}{Itay Hubara}, \bibinfo{person}{Matthieu Courbariaux}, \bibinfo{person}{Daniel Soudry}, \bibinfo{person}{Ran El-Yaniv}, {and} \bibinfo{person}{Yoshua Bengio}.} \bibinfo{year}{2016}\natexlab{}.
\newblock \showarticletitle{Binarized neural networks}. In \bibinfo{booktitle}{\emph{Advances in neural information processing systems}}. \bibinfo{pages}{4107--4115}.
\newblock


\bibitem[Islam and Nirjon(2019)]%
        {islam2019zygarde}
\bibfield{author}{\bibinfo{person}{Bashima Islam} {and} \bibinfo{person}{Shahriar Nirjon}.} \bibinfo{year}{2019}\natexlab{}.
\newblock \showarticletitle{Zygarde: Time-sensitive on-device deep inference and adaptation on intermittently-powered systems}.
\newblock \bibinfo{journal}{\emph{arXiv preprint arXiv:1905.03854}} (\bibinfo{year}{2019}).
\newblock


\bibitem[Kansal and Srivastava(2003)]%
        {kansal2003environmental}
\bibfield{author}{\bibinfo{person}{Aman Kansal} {and} \bibinfo{person}{Mani~B Srivastava}.} \bibinfo{year}{2003}\natexlab{}.
\newblock \showarticletitle{An environmental energy harvesting framework for sensor networks}. In \bibinfo{booktitle}{\emph{Proceedings of the 2003 international symposium on Low power electronics and design}}. \bibinfo{pages}{481--486}.
\newblock


\bibitem[Konda et~al\mbox{.}(2015)]%
        {konda2015zerobiasautoencodersbenefitscoadapting}
\bibfield{author}{\bibinfo{person}{Kishore Konda}, \bibinfo{person}{Roland Memisevic}, {and} \bibinfo{person}{David Krueger}.} \bibinfo{year}{2015}\natexlab{}.
\newblock \bibinfo{title}{Zero-bias autoencoders and the benefits of co-adapting features}.
\newblock
\newblock
\showeprint[arxiv]{1402.3337}~[stat.ML]
\urldef\tempurl%
\url{https://arxiv.org/abs/1402.3337}
\showURL{%
\tempurl}


\bibitem[Krizhevsky(2009)]%
        {krizhevsky2009learning}
\bibfield{author}{\bibinfo{person}{Alex Krizhevsky}.} \bibinfo{year}{2009}\natexlab{}.
\newblock \showarticletitle{Learning multiple layers of features from tiny images}.
\newblock  (\bibinfo{year}{2009}).
\newblock


\bibitem[Kurtz et~al\mbox{.}(2020)]%
        {kurtz2020inducing}
\bibfield{author}{\bibinfo{person}{Mark Kurtz}, \bibinfo{person}{Justin Kopinsky}, \bibinfo{person}{Rati Gelashvili}, \bibinfo{person}{Alexander Matveev}, \bibinfo{person}{John Carr}, \bibinfo{person}{Michael Goin}, \bibinfo{person}{William Leiserson}, \bibinfo{person}{Sage Moore}, \bibinfo{person}{Nir Shavit}, {and} \bibinfo{person}{Dan Alistarh}.} \bibinfo{year}{2020}\natexlab{}.
\newblock \showarticletitle{Inducing and exploiting activation sparsity for fast inference on deep neural networks}. In \bibinfo{booktitle}{\emph{International Conference on Machine Learning}}. PMLR, \bibinfo{pages}{5533--5543}.
\newblock


\bibitem[Kwon et~al\mbox{.}(2020)]%
        {kwon2020structured}
\bibfield{author}{\bibinfo{person}{Se~Jung Kwon}, \bibinfo{person}{Dongsoo Lee}, \bibinfo{person}{Byeongwook Kim}, \bibinfo{person}{Parichay Kapoor}, \bibinfo{person}{Baeseong Park}, {and} \bibinfo{person}{Gu-Yeon Wei}.} \bibinfo{year}{2020}\natexlab{}.
\newblock \showarticletitle{Structured compression by weight encryption for unstructured pruning and quantization}. In \bibinfo{booktitle}{\emph{Proceedings of the IEEE/CVF Conference on Computer Vision and Pattern Recognition}}. \bibinfo{pages}{1909--1918}.
\newblock


\bibitem[Li et~al\mbox{.}(2017)]%
        {li2017pruning}
\bibfield{author}{\bibinfo{person}{Hao Li}, \bibinfo{person}{Asim Kadav}, \bibinfo{person}{Igor Durdanovic}, \bibinfo{person}{Hanan Samet}, {and} \bibinfo{person}{Hans~Peter Graf}.} \bibinfo{year}{2017}\natexlab{}.
\newblock \showarticletitle{Pruning filters for efficient convnets}.
\newblock \bibinfo{journal}{\emph{arXiv preprint arXiv:1608.08710}} (\bibinfo{year}{2017}).
\newblock


\bibitem[Lin et~al\mbox{.}(2017)]%
        {NIPS2017_a51fb975}
\bibfield{author}{\bibinfo{person}{Ji Lin}, \bibinfo{person}{Yongming Rao}, \bibinfo{person}{Jiwen Lu}, {and} \bibinfo{person}{Jie Zhou}.} \bibinfo{year}{2017}\natexlab{}.
\newblock \showarticletitle{Runtime Neural Pruning}. In \bibinfo{booktitle}{\emph{Advances in Neural Information Processing Systems}}, \bibfield{editor}{\bibinfo{person}{I.~Guyon}, \bibinfo{person}{U.~Von Luxburg}, \bibinfo{person}{S.~Bengio}, \bibinfo{person}{H.~Wallach}, \bibinfo{person}{R.~Fergus}, \bibinfo{person}{S.~Vishwanathan}, {and} \bibinfo{person}{R.~Garnett}} (Eds.), Vol.~\bibinfo{volume}{30}. \bibinfo{publisher}{Curran Associates, Inc.}
\newblock
\urldef\tempurl%
\url{https://proceedings.neurips.cc/paper_files/paper/2017/file/a51fb975227d6640e4fe47854476d133-Paper.pdf}
\showURL{%
\tempurl}


\bibitem[Liu et~al\mbox{.}(2019)]%
        {liu2019sparse}
\bibfield{author}{\bibinfo{person}{Baoyuan Liu}, \bibinfo{person}{Min Wang}, \bibinfo{person}{Hassan Foroosh}, \bibinfo{person}{Marshall Tappen}, {and} \bibinfo{person}{Marianna Pensky}.} \bibinfo{year}{2019}\natexlab{}.
\newblock \showarticletitle{Sparse convolutional neural networks}. In \bibinfo{booktitle}{\emph{Proceedings of the IEEE conference on computer vision and pattern recognition}}. \bibinfo{pages}{8060--8068}.
\newblock


\bibitem[Liu et~al\mbox{.}(2017)]%
        {liu2017rethinking}
\bibfield{author}{\bibinfo{person}{Zhuang Liu}, \bibinfo{person}{Mingjie Sun}, \bibinfo{person}{Tinghui Zhou}, \bibinfo{person}{Gao Huang}, {and} \bibinfo{person}{Trevor Darrell}.} \bibinfo{year}{2017}\natexlab{}.
\newblock \showarticletitle{Rethinking the value of network pruning}. In \bibinfo{booktitle}{\emph{International Conference on Learning Representations}}.
\newblock


\bibitem[Liu et~al\mbox{.}(2018)]%
        {liu2018rethinking}
\bibfield{author}{\bibinfo{person}{Zhuang Liu}, \bibinfo{person}{Mingjie Sun}, \bibinfo{person}{Tinghui Zhou}, \bibinfo{person}{Gao Huang}, {and} \bibinfo{person}{Trevor Darrell}.} \bibinfo{year}{2018}\natexlab{}.
\newblock \showarticletitle{Rethinking the value of network pruning}.
\newblock \bibinfo{journal}{\emph{arXiv preprint arXiv:1810.05270}} (\bibinfo{year}{2018}).
\newblock


\bibitem[Luo et~al\mbox{.}(2017)]%
        {luo2017thinet}
\bibfield{author}{\bibinfo{person}{Jian-Hao Luo}, \bibinfo{person}{Jianxin Wu}, {and} \bibinfo{person}{Weiyao Lin}.} \bibinfo{year}{2017}\natexlab{}.
\newblock \showarticletitle{Thinet: A filter level pruning method for deep neural network compression}. In \bibinfo{booktitle}{\emph{Proceedings of the IEEE international conference on computer vision}}. \bibinfo{pages}{5058--5066}.
\newblock


\bibitem[Lyu et~al\mbox{.}(2021)]%
        {lyu2021resource}
\bibfield{author}{\bibinfo{person}{Bo Lyu}, \bibinfo{person}{Hang Yuan}, \bibinfo{person}{Longfei Lu}, {and} \bibinfo{person}{Yunye Zhang}.} \bibinfo{year}{2021}\natexlab{}.
\newblock \showarticletitle{Resource-constrained neural architecture search on edge devices}.
\newblock \bibinfo{journal}{\emph{IEEE Transactions on Network Science and Engineering}} \bibinfo{volume}{9}, \bibinfo{number}{1} (\bibinfo{year}{2021}), \bibinfo{pages}{134--142}.
\newblock


\bibitem[Molchanov et~al\mbox{.}(2017)]%
        {molchanov2017variational}
\bibfield{author}{\bibinfo{person}{Dmitry Molchanov}, \bibinfo{person}{Arsenii Ashukha}, {and} \bibinfo{person}{Dmitry Vetrov}.} \bibinfo{year}{2017}\natexlab{}.
\newblock \showarticletitle{Variational dropout sparsifies deep neural networks}. In \bibinfo{booktitle}{\emph{Proceedings of the 34th International Conference on Machine Learning-Volume 70}}. JMLR. org, \bibinfo{pages}{2498--2507}.
\newblock


\bibitem[Montanari et~al\mbox{.}(2020)]%
        {montanari2020eperceptive}
\bibfield{author}{\bibinfo{person}{Alessandro Montanari}, \bibinfo{person}{Manuja Sharma}, \bibinfo{person}{Dainius Jenkus}, \bibinfo{person}{Mohammed Alloulah}, \bibinfo{person}{Lorena Qendro}, {and} \bibinfo{person}{Fahim Kawsar}.} \bibinfo{year}{2020}\natexlab{}.
\newblock \showarticletitle{ePerceptive: energy reactive embedded intelligence for batteryless sensors}. In \bibinfo{booktitle}{\emph{Proceedings of the 18th Conference on Embedded Networked Sensor Systems}}. \bibinfo{pages}{382--394}.
\newblock


\bibitem[Qi et~al\mbox{.}(2022)]%
        {qi2022fedbkd}
\bibfield{author}{\bibinfo{person}{Peihan Qi}, \bibinfo{person}{Xiaoyu Zhou}, \bibinfo{person}{Yuanlei Ding}, \bibinfo{person}{Zhengyu Zhang}, \bibinfo{person}{Shilian Zheng}, {and} \bibinfo{person}{Zan Li}.} \bibinfo{year}{2022}\natexlab{}.
\newblock \showarticletitle{Fedbkd: Heterogenous federated learning via bidirectional knowledge distillation for modulation classification in iot-edge system}.
\newblock \bibinfo{journal}{\emph{IEEE Journal of Selected Topics in Signal Processing}} \bibinfo{volume}{17}, \bibinfo{number}{1} (\bibinfo{year}{2022}), \bibinfo{pages}{189--204}.
\newblock


\bibitem[Qu et~al\mbox{.}(2020)]%
        {qu2020quantization}
\bibfield{author}{\bibinfo{person}{Xiaoyang Qu}, \bibinfo{person}{Jianzong Wang}, {and} \bibinfo{person}{Jing Xiao}.} \bibinfo{year}{2020}\natexlab{}.
\newblock \showarticletitle{Quantization and knowledge distillation for efficient federated learning on edge devices}. In \bibinfo{booktitle}{\emph{2020 IEEE 22nd International Conference on High Performance Computing and Communications; IEEE 18th International Conference on Smart City; IEEE 6th International Conference on Data Science and Systems (HPCC/SmartCity/DSS)}}. IEEE, \bibinfo{pages}{967--972}.
\newblock


\bibitem[Sadou et~al\mbox{.}(2022)]%
        {no-time-reduction}
\bibfield{author}{\bibinfo{person}{Isma-Ilou Sadou}, \bibinfo{person}{Seyed~Morteza Nabavinejad}, \bibinfo{person}{Zhonghai Lu}, {and} \bibinfo{person}{Masoumeh Ebrahimi}.} \bibinfo{year}{2022}\natexlab{}.
\newblock \showarticletitle{Inference Time Reduction of Deep Neural Networks on Embedded Devices: A Case Study}. In \bibinfo{booktitle}{\emph{2022 25th Euromicro Conference on Digital System Design (DSD)}}. \bibinfo{pages}{205--213}.
\newblock
\urldef\tempurl%
\url{https://doi.org/10.1109/DSD57027.2022.00036}
\showDOI{\tempurl}


\bibitem[Shen et~al\mbox{.}(2022)]%
        {shen2022prune}
\bibfield{author}{\bibinfo{person}{Maying Shen}, \bibinfo{person}{Pavlo Molchanov}, \bibinfo{person}{Hongxu Yin}, {and} \bibinfo{person}{Jose~M Alvarez}.} \bibinfo{year}{2022}\natexlab{}.
\newblock \showarticletitle{When to prune? a policy towards early structural pruning}. In \bibinfo{booktitle}{\emph{Proceedings of the IEEE/CVF Conference on Computer Vision and Pattern Recognition}}. \bibinfo{pages}{12247--12256}.
\newblock


\bibitem[Teerapittayanon et~al\mbox{.}(2016)]%
        {teerapittayanon2016branchynet}
\bibfield{author}{\bibinfo{person}{Surat Teerapittayanon}, \bibinfo{person}{Bradley McDanel}, {and} \bibinfo{person}{Hsiang-Tsung Kung}.} \bibinfo{year}{2016}\natexlab{}.
\newblock \showarticletitle{Branchynet: Fast inference via early exiting from deep neural networks}. In \bibinfo{booktitle}{\emph{2016 23rd international conference on pattern recognition (ICPR)}}. IEEE, \bibinfo{pages}{2464--2469}.
\newblock


\bibitem[Thramboulidis et~al\mbox{.}(2007)]%
        {thramboulidis2007soa}
\bibfield{author}{\bibinfo{person}{Kleanthis~C Thramboulidis}, \bibinfo{person}{G Doukas}, {and} \bibinfo{person}{G Koumoutsos}.} \bibinfo{year}{2007}\natexlab{}.
\newblock \showarticletitle{A SOA-based embedded systems development environment for industrial automation}.
\newblock \bibinfo{journal}{\emph{EURASIP Journal on Embedded Systems}}  \bibinfo{volume}{2008} (\bibinfo{year}{2007}), \bibinfo{pages}{1--15}.
\newblock


\bibitem[Warden(2018)]%
        {warden2018speech}
\bibfield{author}{\bibinfo{person}{Pete Warden}.} \bibinfo{year}{2018}\natexlab{}.
\newblock \showarticletitle{Speech commands: A dataset for limited-vocabulary speech recognition}.
\newblock \bibinfo{journal}{\emph{arXiv preprint arXiv:1804.03209}} (\bibinfo{year}{2018}).
\newblock


\bibitem[Wen et~al\mbox{.}(2016)]%
        {wen2016learning}
\bibfield{author}{\bibinfo{person}{Wei Wen}, \bibinfo{person}{Chunpeng Wu}, \bibinfo{person}{Yandan Wang}, \bibinfo{person}{Yiran Chen}, {and} \bibinfo{person}{Hai Li}.} \bibinfo{year}{2016}\natexlab{}.
\newblock \showarticletitle{Learning structured sparsity in deep neural networks}. In \bibinfo{booktitle}{\emph{Advances in neural information processing systems}}. \bibinfo{publisher}{NIPS}, \bibinfo{pages}{2074--2082}.
\newblock


\bibitem[Wu et~al\mbox{.}(2016)]%
        {wu2016quantized}
\bibfield{author}{\bibinfo{person}{Jiaxiang Wu}, \bibinfo{person}{Cong Leng}, \bibinfo{person}{Yuhang Wang}, \bibinfo{person}{Qinghao Hu}, {and} \bibinfo{person}{Jian Cheng}.} \bibinfo{year}{2016}\natexlab{}.
\newblock \showarticletitle{Quantized convolutional neural networks for mobile devices}. In \bibinfo{booktitle}{\emph{Proceedings of the IEEE conference on computer vision and pattern recognition}}. \bibinfo{pages}{4820--4828}.
\newblock


\bibitem[Yang et~al\mbox{.}(2018)]%
        {yang2018netadapt}
\bibfield{author}{\bibinfo{person}{Tien-Ju Yang}, \bibinfo{person}{Andrew Howard}, \bibinfo{person}{Bo Chen}, \bibinfo{person}{Xiao Zhang}, \bibinfo{person}{Alec Go}, \bibinfo{person}{Mark Sandler}, \bibinfo{person}{Vivienne Sze}, {and} \bibinfo{person}{Hartwig Adam}.} \bibinfo{year}{2018}\natexlab{}.
\newblock \showarticletitle{Netadapt: Platform-aware neural network adaptation for mobile applications}. In \bibinfo{booktitle}{\emph{Proceedings of the European conference on computer vision (ECCV)}}. \bibinfo{pages}{285--300}.
\newblock


\bibitem[Ye et~al\mbox{.}(2020)]%
        {10.1007/978-3-030-63823-8_58}
\bibfield{author}{\bibinfo{person}{Yun Ye}, \bibinfo{person}{Ganmei You}, \bibinfo{person}{Jong-Kae Fwu}, \bibinfo{person}{Xia Zhu}, \bibinfo{person}{Qing Yang}, {and} \bibinfo{person}{Yuan Zhu}.} \bibinfo{year}{2020}\natexlab{}.
\newblock \showarticletitle{Channel Pruning via Optimal Thresholding}. In \bibinfo{booktitle}{\emph{Neural Information Processing}}, \bibfield{editor}{\bibinfo{person}{Haiqin Yang}, \bibinfo{person}{Kitsuchart Pasupa}, \bibinfo{person}{Andrew Chi-Sing Leung}, \bibinfo{person}{James~T. Kwok}, \bibinfo{person}{Jonathan~H. Chan}, {and} \bibinfo{person}{Irwin King}} (Eds.). \bibinfo{publisher}{Springer International Publishing}, \bibinfo{address}{Cham}, \bibinfo{pages}{508--516}.
\newblock
\showISBNx{978-3-030-63823-8}


\bibitem[Zhang et~al\mbox{.}(2022)]%
        {WIDAR-dataset}
\bibfield{author}{\bibinfo{person}{Y. Zhang}, \bibinfo{person}{Y. Zheng}, \bibinfo{person}{K. Qian}, \bibinfo{person}{G. Zhang}, \bibinfo{person}{Y. Liu}, \bibinfo{person}{C. Wu}, {and} \bibinfo{person}{Z. Yang}.} \bibinfo{year}{2022}\natexlab{}.
\newblock \showarticletitle{Widar3.0: Zero-Effort Cross-Domain Gesture Recognition With Wi-Fi}.
\newblock \bibinfo{journal}{\emph{IEEE Transactions on Pattern Analysis and Machine Intelligence}} \bibinfo{volume}{44}, \bibinfo{number}{11} (\bibinfo{date}{nov} \bibinfo{year}{2022}), \bibinfo{pages}{8671--8688}.
\newblock
\showISSN{1939-3539}
\urldef\tempurl%
\url{https://doi.org/10.1109/TPAMI.2021.3105387}
\showDOI{\tempurl}


\bibitem[Zhao et~al\mbox{.}(2023)]%
        {zhao2023automaticattentionpruningimproving}
\bibfield{author}{\bibinfo{person}{Kaiqi Zhao}, \bibinfo{person}{Animesh Jain}, {and} \bibinfo{person}{Ming Zhao}.} \bibinfo{year}{2023}\natexlab{}.
\newblock \bibinfo{title}{Automatic Attention Pruning: Improving and Automating Model Pruning using Attentions}.
\newblock
\newblock
\showeprint[arxiv]{2303.08595}~[cs.LG]
\urldef\tempurl%
\url{https://arxiv.org/abs/2303.08595}
\showURL{%
\tempurl}


\bibitem[Zhu and Gupta(2017)]%
        {zhu2017to}
\bibfield{author}{\bibinfo{person}{Michael Zhu} {and} \bibinfo{person}{Suyog Gupta}.} \bibinfo{year}{2017}\natexlab{}.
\newblock \showarticletitle{To prune, or not to prune: Exploring the efficacy of pruning for model compression}.
\newblock \bibinfo{journal}{\emph{arXiv preprint arXiv:1710.01878}} (\bibinfo{year}{2017}).
\newblock


\end{thebibliography}


\end{document}